%% file: paper.tex
\documentclass[letterpaper]{article} 
\usepackage{aaai2026}  
\usepackage{times}  
\usepackage{helvet}  
\usepackage{courier}  
\usepackage[hyphens]{url}  
\usepackage{graphicx} 

\usepackage{amsmath}
\usepackage{amsfonts}
\usepackage{amssymb}
\usepackage[linesnumbered,ruled,vlined]{algorithm2e}
\usepackage{booktabs}
\usepackage{makecell}
\usepackage{comment}
\usepackage{enumitem}
\usepackage{subfigure}

\usepackage{mathtools}

\usepackage{bm}
\usepackage{physics}   
\usepackage{xcolor} 
\SetKwInput{KwIn}{Input}
\SetKwInput{KwOut}{Output}

\usepackage{natbib}  
\usepackage{caption} 
\usepackage{newfloat}
\usepackage{listings}
\DeclareCaptionStyle{ruled}{labelfont=normalfont,labelsep=colon,strut=off} 
\title{Affordance-Guided Coarse-to-Fine Exploration for Base Placement in Open-Vocabulary Mobile Manipulation}
\author{
    Tzu-Jung Lin\textsuperscript{\rm 1},
    Jia-Fong Yeh\textsuperscript{\rm 1},
    Hung-Ting Su\textsuperscript{\rm 1},
    Chung-Yi Lin\textsuperscript{\rm 1},
    Yi-Ting Chen\textsuperscript{\rm 2},
    Winston H. Hsu\textsuperscript{\rm 1}
}
\affiliations{
    \textsuperscript{\rm 1}National Taiwan University\\
    \textsuperscript{\rm 2}National Yang Ming Chiao Tung University

}

\usepackage{bibentry}

\newif\ifshowappendix
\showappendixtrue

\begin{document}

\maketitle

\begin{abstract}
In open-vocabulary mobile manipulation (OVMM), task success often hinges on the selection of an appropriate base placement for the robot. Existing approaches typically navigate to proximity-based regions without considering affordances, resulting in frequent manipulation failures. We propose Affordance-Guided Coarse-to-Fine Exploration, a zero-shot framework for base placement that integrates semantic understanding from vision-language models (VLMs) with geometric feasibility through an iterative optimization process. Our method constructs cross-modal representations, namely \textit{Affordance RGB} and \textit{Obstacle Map+}, to align semantics with spatial context. This enables reasoning that extends beyond the egocentric limitations of RGB perception. To ensure interaction is guided by task-relevant affordances, we leverage coarse semantic priors from VLMs to guide the search toward task-relevant regions and refine placements with geometric constraints, thereby reducing the risk of convergence to local optima. Evaluated on five diverse open-vocabulary mobile manipulation tasks, our system achieves an 85\% success rate, significantly outperforming classical geometric planners and VLM-based methods. This demonstrates the promise of affordance-aware and multimodal reasoning for generalizable, instruction-conditioned planning in OVMM.
\end{abstract}

\begin{figure*}[htbp]
    \centering
    \includegraphics[width=0.99\linewidth]{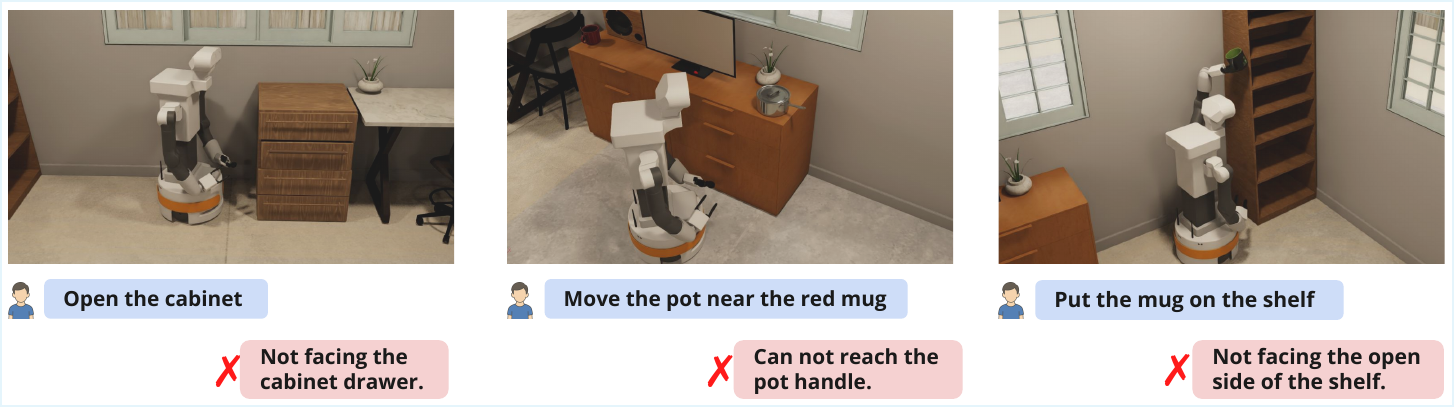}
    \caption{
Examples of failure cases caused by base placements without affordance awareness.  Left: The robot cannot open the cabinet because it is not facing the drawer. Middle: The robot cannot grasp the pot handle due to misalignment. Right: The robot fails to place the mug on the shelf as it is not facing the open side.
These failures arise from a lack of joint reasoning over task intent and geometric feasibility, leading to semantically misaligned placements that prevent successful manipulation.}
    \label{fig:demo}
\end{figure*}

\section{Introduction}

In open-vocabulary mobile manipulation (OVMM), selecting an appropriate base placement is critical for successful task execution. However, prior navigation systems \cite{yenamandra2023homerobot, huang2023ok, qiu2024open, tan2025languageconditionedopenvocabularymobilemanipulation} often treat the task as complete once the robot reaches a location near the target. In practice, mere proximity does not guarantee effective manipulation. Specifically, determining the base placement presents two key challenges. \textbf{First}, the robot must \emph{reason jointly about geometric feasibility and semantic intent}. It needs to identify a collision-free location that maintains appropriate distance and aligns with task-relevant affordances. For example, to open a cabinet, the robot must position itself in front of the cabinet with enough clearance for effective interaction. \textbf{Second}, the robot must \emph{reason globally despite limited perceptual input}. A narrow field of view restricts spatial awareness and may cause the robot to overlook suitable placements. For instance, if the area in front of the cabinet is not visible in the RGB image, the robot cannot evaluate it as a potential base location.

Existing approaches struggle to address the aforementioned challenges. In many OVMM \cite{huang2023ok, qiu2024open} or navigation \cite{singh2023clipfields, huang2023visual} works, once the target object’s location is known, classical path planners such as A* \cite{hart1968formal} and RRT* \cite{karaman2011sampling} are used to navigate the robot close to the object. These planners rely solely on geometric heuristics and lack task-level semantic understanding. As a result, they often produce placements that fail to account for task affordances, as illustrated in Figure \ref{fig:demo}. Conversely, semantic methods that leverage vision-language models (VLMs) can infer high-level task intent  \cite{nasiriany2024pivot, sathyamoorthy2024convoicontextawarenavigationusing}, but typically overlook geometric feasibility and reachability constraints. Furthermore, these methods often rely on a single RGB image, which limits their ability to reason about occluded or unseen areas.

In this paper, we propose a novel zero-shot framework, \textit{Affordance-Guided Coarse-to-Fine Exploration}, which integrates both semantic reasoning and geometric understanding for effective base placement, especially under limited field of view. Our approach introduces two key innovations:
(1) \textbf{Cross-modal representations}, specifically \textit{Affordance RGB} and \textit{Obstacle Map+}, which align task intent with spatial layout while mitigating perceptual limitations caused by occluded or unseen regions.
(2) \textbf{A coarse-to-fine optimization process} that begins with sampling in semantically guided regions and iteratively refines the search toward geometrically feasible placements. This enables the robot to satisfy both semantic and geometric requirements.

Our system requires no task-specific supervision and operates solely on natural language instructions, RGB-D images, and an obstacle map. It generalizes across a wide range of OVMM tasks with varying spatial and semantic demands. Experiments show that our method achieves an overall success rate of 85\%, significantly outperforming classical and semantic-only baselines. These results underscore the effectiveness of our key component, i.e. affordance-aware and multimodal reasoning, in enabling instruction-driven base placement for open-vocabulary mobile manipulation.

\noindent \textbf{Our main contributions are:}
\begin{itemize}[leftmargin=1em, noitemsep]
  \item We identify base placement as a critical challenge in OVMM, driven by the need to reason about both semantics and geometry under limited perceptual input.
  \item We propose a coarse-to-fine strategy that unifies semantic and geometric cues via \textit{Affordance RGB} and \textit{Obstacle Map+}, addressing limited field of view and overcoming the limitations of prior single-focus methods.
  \item We develop a generalizable zero-shot system that achieves 85\% success across tasks, outperforming prior classical and semantic-only approaches.
\end{itemize}

\section{Related Work}
\paragraph{Open-Vocabulary Navigation and Manipulation} Recent vision-language models (VLMs) \cite{openai2023gpt4,team2023gemini,liu2023visualinstructiontuning} have shown strong potential in open-vocabulary navigation and manipulation. Some systems build semantic maps using VLM features to enable language-guided planning. For example, FindAnything~\cite{laina2025findanythingopenvocabularyobjectcentricmapping} constructs object-centric submaps, while CLIP-Fields \cite{singh2023clipfields} and VLMaps \cite{huang2023visual} embed vision-language features into 3D environments. CLIP-Fields\cite{singh2023clipfields} also projects weakly-supervised features to generate semantic maps. Methods like USA-Nets \cite{bolte2023usa} and GOAT \cite{chang2023goatthing} locate targets using language-image matching, then guide navigation using geometric planners like A*. These techniques extend to open-vocabulary mobile manipulation (OVMM). OK-Robot~\cite{huang2023ok} builds a VoxelMap from RGB-D scans and uses CLIP~\cite{radford2021learningtransferablevisualmodels}, SAM~\cite{kirillov2023segment}, and OWL-ViT~\cite{minderer2022simpleopenvocabularyobjectdetection} for grounding and segmentation. It then queries VLMs to plan collision-free paths. COME-robot~\cite{zhi2025closedloopopenvocabularymobilemanipulation} incorporates GPT-4V for task planning and failure recovery. Our work differs by focusing specifically on selecting semantically meaningful and physically feasible base placements.

\paragraph{Visual Prompting for Robotics} Visual prompting is an emerging tool in robotic VLMs. While Set-of-Mark Prompting (SoM) \cite{yang2023setofmarkpromptingunleashesextraordinary} was originally developed for visual grounding tasks, its underlying techniques have been extended to symbolic prompting in systems like PIVOT \cite{nasiriany2024pivot}, which frames spatial tasks as iterative visual question answering and refines VLM predictions through repeated annotation and selection. CoPa \cite{huang2024copageneralroboticmanipulation}, ReKep \cite{huang2024rekep}, and MOKA \cite{zhang2025momakitchen100kbenchmarkaffordancegrounded} use visual prompts to infer keypoints or constraints for manipulation. KAGI \cite{lee2025affordanceguidedreinforcementlearningvisual} utilizes keypoints to define dense rewards in reinforcement learning, while CoNVOI \cite{sathyamoorthy2024convoicontextawarenavigationusing} applies region-level prompting for navigation. However, these methods rely on single-view RGB inputs, limiting their ability to reason about occluded or unseen regions. In contrast, our approach applies visual prompting directly to obstacle maps, enabling global inference beyond the current view.

\paragraph{Recent Advances in Base Placement}
Recent work addresses base placement by integrating perception and planning. MoMa-Pos \cite{shao2024momaposefficientobjectkinematicawarebase} optimizes base placement by modeling task-relevant articulated objects using frontal-view images, but requires object-specific modeling and lacks generalization to novel categories.
 MoMa-Kitchen\cite{zhang2025momakitchen100kbenchmarkaffordancegrounded} introduces a large egocentric dataset for affordance prediction, but its performance may be constrained by the limited field of view inherent in egocentric perspectives. Navi2Gaze \cite{zhu2024navi2gazeleveragingfoundationmodels} uses VLMs to choose base locations aligned with object orientation but does not reason about reachability or execution feasibility. Our approach applies affordance prompting directly on obstacle maps, enabling generalization to unseen scenes and occluded configurations.

\section{Preliminaries}

\subsection{Pipeline Overview}

We consider an open-vocabulary mobile manipulation setting in which a robot is given a high-level natural language instruction $\ell$. The instruction is parsed by GPT-4~\cite{openai2023gpt4}, a large language model, into a sequence of sub-instructions. Each sub-instruction is represented as a tuple $(t, \tilde{\ell})$, where $t$ denotes the name of the referenced object and $\tilde{\ell}$ is the corresponding sub-task instruction. For example, the instruction ``Put the mug on the shelf'' is parsed as: [(``mug'', ``pick up the mug.''), (``shelf'', ``put the mug on the shelf.'')].

To focus on base placement, we assume that the 2D position of the target object $t$, denoted $\mathbf{p}_t \in \mathbb{R}^2$, is directly provided by the simulator. This allows the system to operate given a known object location and a global obstacle map $M_{\text{global}}$, without addressing open-world object grounding.

Given a grounded sub-instruction $(t, \tilde{\ell})$ and access to $\mathbf{p}_t$ and $M_{\text{global}}$, the robot executes the following three-stage pipeline:

\begin{enumerate}
    \item \textbf{Navigation:} Navigate to a coarse waypoint near the object’s known position (typically within 1.5 meters) using a path planner and orient the robot to face the object.
    \item \textbf{Base Placement Selection:} Select an optimized base placement for manipulation by reasoning over local geometric and semantic cues, and move to the selected placement.
    \item \textbf{Manipulation:} Execute the sub-task $\tilde{\ell}$ using a predefined manipulation primitive such as pick, place, or open.
\end{enumerate}

To support this pipeline, the robot maintains a global 2D occupancy grid map $M_{\text{global}}$ and dynamically derives a local egocentric map $M_{\text{local}}$ at runtime. 

\subsection{Problem Statement}

After coarse navigation, the robot has access to:
\begin{itemize}[leftmargin=1em, itemsep=0.02em]
    \item A target object name $t$ and sub-instruction $\tilde{\ell}$ parsed from the high-level command $\ell$,
    \item An obstacle map $M_{\text{local}}$ indicating non-navigable regions,
    \item An RGB $I$ and a depth image $D$ from onboard sensors.
\end{itemize}
The goal is to select a base placement $\mathbf{x} \in \mathcal{X}_{\text{free}}$, where $\mathcal{X}_{\text{free}}$ is the set of collision-free placements with sufficient clearance from obstacles:
\[
\mathcal{X}_{\text{free}} = \left\{ \mathbf{x} \;\middle|\; \mathbf{x} \text{ is collision-free and } \text{dist}(\mathbf{x}, \mathcal{O}) \geq 0.4\,\text{m} \right\}
\]
with $\mathcal{O} \subseteq M_{\text{local}}$ denoting obstacle regions.
We then solve:
\[
\mathbf{x}^* = \arg\max_{\mathbf{x} \in \mathcal{X}_{\text{free}}} \mathbb{P} \left[ \text{success} \mid \mathbf{x}, I, D, \tilde{\ell}, t, M_{\text{local}} \right]
\]
A trial is successful if the robot can reach a valid end-effector pose $g^* \in \mathcal{G}_t$, defined over the affordance region of object $t$, and execute the manipulation. This includes:
\begin{itemize}[leftmargin=1em, itemsep=0.02em]
    \item Solving IK to find joint configuration $\theta$ for $g^*$,
    \item Moving the arm to $\theta$ without collisions,
    \item Executing the manipulation primitive (e.g., pick, open),
    \item Verifying the expected physical outcome.
\end{itemize}

\begin{figure*}[htbp]
    \centering
    \includegraphics[width=0.92\linewidth]{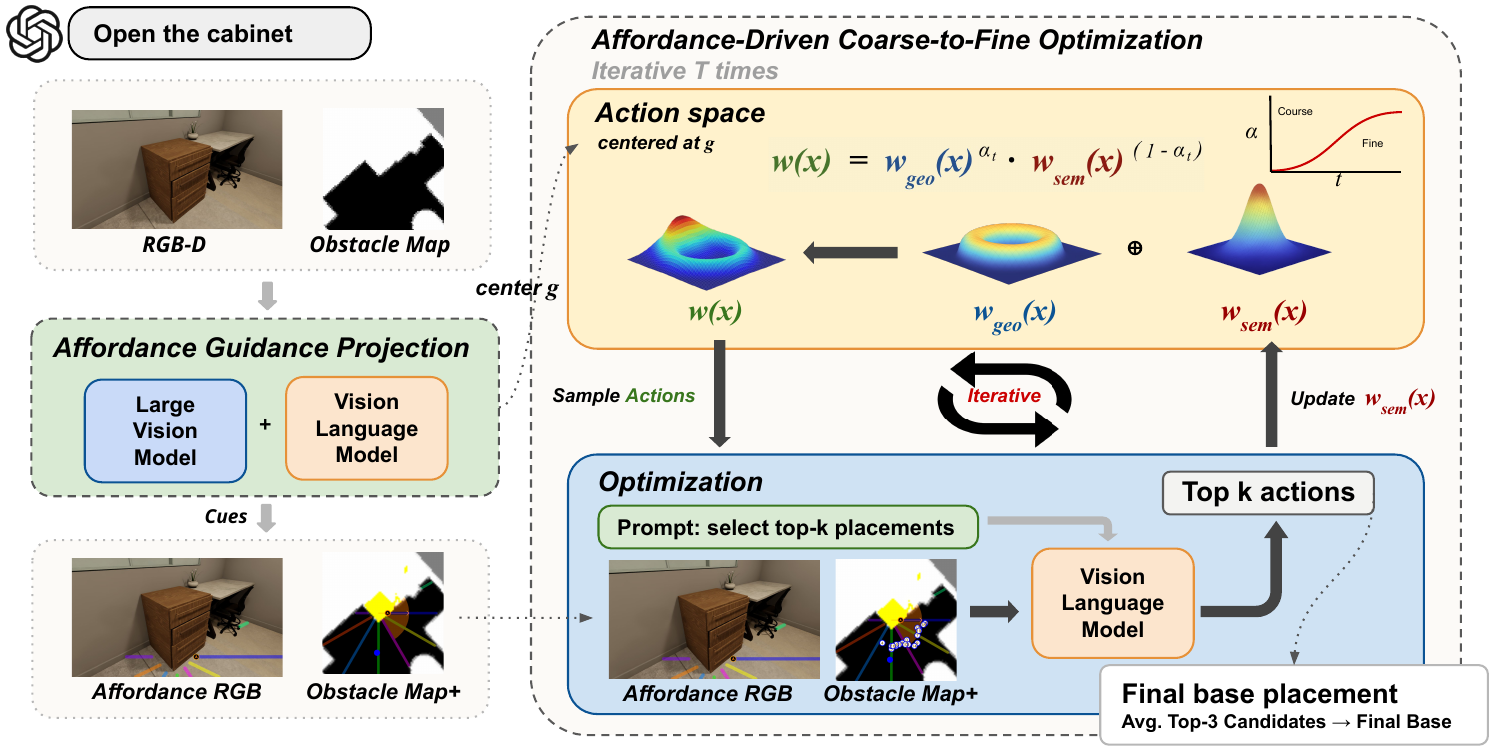}
    \caption{
    Affordance-Guided Coarse-to-Fine Exploration.
    The method comprises two key components. (1) To overcome the limitations of single-view perception, it applies \textit{Affordance Guidance Projection}, which uses semantic cues to generate Affordance RGB and Obstacle Map+ from RGB and obstacle maps, enabling global semantic reasoning. (2) To identify base placements that satisfy both semantic relevance and geometric feasibility, it introduces \textit{Affordance-Driven Coarse-to-Fine Optimization}, which leverages the coarse, high-level nature of VLM outputs to explore semantically appropriate regions. As the process iterates, geometric constraints are gradually emphasized, guiding the search toward executable base placements.
    }
    \label{fig:method}
\end{figure*}

\section{Method}

We present \textbf{Affordance-Guided Coarse-to-Fine Exploration}, a framework for selecting base placements that are both semantically meaningful and geometrically feasible. The key idea is to leverage large vision models (LVMs) and vision-language models (VLMs) for high-level semantic guidance, while using iterative optimization to achieve spatial precision. Our approach consists of two stages: (1) \textbf{Affordance Guidance Projection}, which extracts affordance cues from perception and projects them onto a 2D obstacle map; and (2) \textbf{Affordance-Driven Coarse-to-Fine Optimization}, which refines candidate base placements through probabilistic sampling and VLM feedback. An overview of the method is depicted in Figure~\ref{fig:method}.

\subsection{Affordance Guidance Projection}

To semantically guide robot base planning beyond RGB perception, we introduce a cross-modal projection mechanism that aligns visual-semantic information with spatial geometric maps. While obstacle maps provide valuable collision-aware context, they inherently lack semantic richness. Conversely, VLMs can perform language-grounded affordance reasoning but are limited to RGB inputs and lack explicit spatial awareness. To bridge this modality gap, we extract affordance cues from RGB images using Grounded SAM~\cite{ren2024groundedsamassemblingopenworld} for object segmentation and GPT-4o~\cite{openai2023gpt4} for language-conditioned reasoning. These cues are then projected onto the robot’s 2D obstacle map, enabling consistent semantic alignment between visual and spatial representations. This cross-modal grounding facilitates more informed base placement decisions. 
 To operationalize this idea, we construct two complementary multimodal representations:

\begin{itemize}

    \item \textbf{Affordance RGB ($I_{\text{aff}}$):} An RGB image overlaid with:
    \begin{enumerate}
        \item 12 directional arrows with distinct colors, evenly spaced at $30^\circ$ intervals around the object.
        \item One arrow labeled “A” to indicate the coarse affordance direction suggested by the VLM.
    \end{enumerate}

    \item \textbf{Obstacle Map+ ($M_{\text{local}}^+$):} A top-down spatial map augmented with: 
    \begin{enumerate}
        \item The segmented target object footprint $\mathcal{R}_t$.
        \item The robot's current base location.
        \item A fan-shaped affordance region $\mathcal{F}_t$ centered on the selected direction (labeled ``A''), spanning $\pm 60^\circ$ from that direction.
        \item All 12 directional arrows are rendered with colors consistently matched to those in the RGB image, ensuring cross-modal semantic alignment.
    \end{enumerate}

\end{itemize}

\subsection{Affordance-Driven Coarse-to-Fine Optimization}

To complete a manipulation task such as opening the dishwasher, the robot must select a base placement from which the relevant part of the object is physically reachable to enable successful execution. Relying solely on geometric reasoning may result in infeasible placements, while depending only on vision-language models (VLMs) may produce semantically appropriate but unreachable ones.

To address this, we propose an optimization method that begins from a task-specific affordance keypoint \( \mathbf{g} \) (e.g., a handle) and searches the surrounding region for functionally viable base placements. The search is guided by two criteria: (1) the placement must align with task semantics, and (2)  it must lie within the robot’s reachable workspace. 

Our approach adopts a coarse-to-fine strategy that is realized through an iterative scoring mechanism. Specifically, candidate placements are sampled from a truncated Gaussian distribution centered at \( \mathbf{g} \), and each candidate is assigned a score based on a combination of semantic relevance and spatial proximity. As the optimization progresses, the weighting of the scoring function is gradually shifted from semantic alignment to geometric precision. This enables the robot to explore task-relevant placements early on and converge toward physically executable ones in later iterations. 

We describe the selection of the affordance keypoint \( \mathbf{g} \) using a method similar to~\cite{huang2024rekep}, followed by details of the sampling and optimization procedure.

\subsubsection{Affordance Point Selection}
To identify the affordance keypoint \( \mathbf{g} \), we first extract visual representations from the entire RGB image using DINOv2~\cite{oquab2024dinov2learningrobustvisual}. Then, Grounded SAM~\cite{ren2024groundedsamassemblingopenworld} is used to segment the target object based on the task instruction. From the extracted DINOv2 features within the segmented region, we apply k-means clustering (with cosine similarity) to produce spatially diverse candidate keypoints. Cluster centroids are projected onto the image, rendered, and annotated. Finally, given the task sub-instruction and annotated image, GPT-4o~\cite{openai2023gpt4} selects the keypoint most semantically aligned with the intended interaction. The selected keypoint \( \mathbf{g} \) serves as the sampling center for generating candidate base placements. 

\begin{table*}[htbp]
\centering
\begin{tabular}{lcccccc}
\toprule
\textbf{Method} &
\begin{tabular}[c]{@{}c@{}}\textbf{Throw the} \\ \textbf{Can into Trash}\end{tabular} &
\begin{tabular}[c]{@{}c@{}}\textbf{Move Pot} \\ \textbf{Near Red Mug}\end{tabular} &
\begin{tabular}[c]{@{}c@{}}\textbf{Put Mug} \\ \textbf{on Shelf}\end{tabular} &
\begin{tabular}[c]{@{}c@{}}\textbf{Open} \\ \textbf{Cabinet}\end{tabular} &
\begin{tabular}[c]{@{}c@{}}\textbf{Open} \\ \textbf{Dishwasher}\end{tabular} &
\textbf{Total Success} \\
\midrule
Object 
Center + A* & \textbf{20/20} & 9/20 & 8/20 & 5/20 & 5/20 & 47\% \\
Object 
Center + RRT* & 19/20 & 8/20 & 3/20 & 10/20 & 10/20 & 50\% \\
Affordance Point + A* & 16/20 & 10/20 & 13/20 & 10/20 & 9/20 & 58\% \\
Affordance Point + RRT* & 18/20 & 10/20 & 10/20 & 11/20 & 12/20 & 61\% \\
Pivot ($I$) & 0/20 & 2/20 & 1/20 & \textbf{17/20} & 6/20 & 26\% \\
Pivot ($M_{\text{local}}^+$, $I_{\text{aff}}$) & 2/20 & 3/20 & 2/20 & 10/20 & 6/20 & 23\% \\
\textbf{Our method} & 17/20 & \textbf{18/20} & \textbf{17/20} & 16/20 & \textbf{17/20} & \textbf{85\%} \\
\bottomrule
\end{tabular}
\caption{Success Rates Across Five Mobile Manipulation Tasks}
\label{tab:task_success}
\end{table*}

\subsubsection{Iterative Optimization}

We perform the base selection procedure over \( T \) iterative steps. Each iteration consists of three main stages: (1) scoring, (2) sampling, and (3) refinement. 

\paragraph{(1) Scoring.}
At each iteration \( t \), we sample a set of candidate base placements \( \{x_i\}_{i=1}^N \) from a Gaussian distribution centered at the predicted affordance point \( \mathbf{g} \):
\[
x_i \sim \mathcal{N}(\mathbf{g}, \sigma_{\text{sample}}^2 I), \quad \text{s.t.} \quad \|x_i - \mathbf{g}\| \leq r_{\text{max}}, \quad x_i \in \mathcal{X}_{\text{free}}.
\]
Sampling is truncated at a fixed radius \( r_{\text{max}} \) and restricted to collision-free regions \( \mathcal{X}_{\text{free}} \). 

Each candidate is assigned a composite score \( w(x) \) that balances geometric and semantic relevance:
\begin{equation}
w(x) = w_{\text{geo}}(x)^{\alpha_t} \cdot w_{\text{sem}}(x)^{1 - \alpha_t},
\label{eq:w(x)}
\end{equation}
where \( \alpha_t \in [0, 1] \) is a time-dependent weighting coefficient that shifts gradually from semantic alignment toward geometric precision. 

The \emph{geometric term} encourages sampling at a preferred distance \( r^* \) from \( \mathbf{g} \):
\[
w_{\text{geo}}(x) = \Phi(\|x - \mathbf{g}\|; r^*, \sigma_g),
\]
where \( \Phi(d; \mu, \sigma) \) measures the cumulative probability mass within a margin \( \delta \):
\[
\Phi(d; \mu, \sigma) = \text{CDF}_{\mu, \sigma}(d + \delta) - \text{CDF}_{\mu, \sigma}(d - \delta).
\]
Here, \( \text{CDF}_{\mu, \sigma}(x) \) denotes the cumulative distribution function of a Gaussian distribution with mean \( \mu \) and standard deviation \( \sigma \).

The \emph{semantic term} encourages alignment with an evolving semantic center \( \mu_t \):
\[
w_{\text{sem}}(x) =
\begin{cases}
\Phi(\|x - \mu_t\|; 0, \sigma_s), & \text{if } \mu_t \text{ is defined}, \\
1, & \text{otherwise}.
\end{cases}
\label{eq:w_sem}
\]

To implement a smooth transition from semantic exploration to geometric refinement, we use a sigmoid schedule:
\begin{equation}
\alpha_t = \alpha_{\text{max}} \left(1 + e^{-\gamma(t - T/2)}\right)^{-1},
\label{eq:sigmoid}
\end{equation}
where \( \alpha_{\text{max}} \) is the maximum geometric weight, \( \gamma \) controls the steepness, and \( T/2 \) is the inflection point. This design prioritizes VLM-based semantic reasoning in early stages—when coarse, high-level decisions are most useful—and gradually shifts toward precise spatial optimization in later iterations.

\paragraph{(2) Sampling.}
The normalized weights define a discrete probability distribution:
\begin{equation}
p(x_i) = \frac{w(x_i)}{\sum_{j=1}^N w(x_j)}.
\label{eq:prob}
\end{equation}
We sample \( N_{\text{sample}} \) candidates from this distribution. Each sampled location is projected onto the local obstacle map \( M_{\text{local}}^+ \) and assigned a unique index for identification. The indexed map, along with the Affordance RGB image \( I_{\text{aff}} \) and the sub-instruction \( \tilde{\ell} \), is submitted to the VLM as a joint multimodal prompt for semantic ranking.

\paragraph{(3) Refinement.}
For iterations \( t < T \), the VLM ranks the sampled candidates and returns the top-$k$ semantically relevant points, denoted as \( \{x^{(1)}, \dots, x^{(k)}\} \). The semantic center is then updated as \( \mu_t = \frac{1}{k} \sum_{i=1}^k x^{(i)} \).
To encourage convergence, we progressively reduce \( \sigma_s \) across iterations. 

At the final iteration (\( t = T \)), we still perform VLM-based ranking, but omit the semantic center update. Instead, we directly take the top-5 VLM-ranked candidates, remove the two furthest from their mean, and compute the final base placement by averaging the remaining three.

\begin{table*}[htbp]
\centering

\begin{tabular}{lcccccc}
\toprule
\textbf{Setting} &
\begin{tabular}[c]{@{}c@{}}\textbf{Throw the} \\ \textbf{Can into Trash}\end{tabular} &
\begin{tabular}[c]{@{}c@{}}\textbf{Move Pot} \\ \textbf{Near Red Mug}\end{tabular} &
\begin{tabular}[c]{@{}c@{}}\textbf{Put Mug} \\ \textbf{on Shelf}\end{tabular} &
\begin{tabular}[c]{@{}c@{}}\textbf{Open} \\ \textbf{Cabinet}\end{tabular} &
\begin{tabular}[c]{@{}c@{}}\textbf{Open} \\ \textbf{Dishwasher}\end{tabular} &
\textbf{Total Success} \\
\midrule

$\alpha = 0$ & 10/20 & 11/20 & 5/20 & 6/20 & 11/20 & 43\% \\
$\alpha = 0.5$ & 18/20 & 14/20 & 14/20 & 15/20 & 15/20 & 76\% \\
$\alpha = 1$ & \textbf{20/20} & 14/20 & \textbf{18/20} & 12/20 & 16/20 & 79\% \\

\textbf{Increasing $\alpha_t$ (Ours)} & 17/20 & \textbf{18/20} & 17/20 & \textbf{16/20} & \textbf{17/20} & \textbf{85\%} \\
\bottomrule
\end{tabular}
\caption{Success Rates Under Different $\alpha$ Settings}
\label{tab:alpha_ablation}
\end{table*}

\section{Experiments}

We evaluate our approach in a simulated mobile manipulation environment, focusing on base placement for diverse open-vocabulary tasks. This section details the experimental setup, task definitions, and comparative performance analysis, including ablation studies to assess the contribution of each component.

\subsection{Experimental Setup}

All experiments are conducted in NVIDIA Isaac Sim using the TIAGo++ mobile manipulation platform, with the left arm employed throughout this study. The robot is equipped with a differential-drive base and a 7-DOF manipulator, which is controlled via inverse kinematics (IK) through the left arm–torso kinematic chain. Perception is enabled by a head-mounted RGB-D camera with a resolution of 1280×720. Known intrinsic and extrinsic parameters are used to project observations into both robot-centric and world coordinate frames.

\subsection{Task Description}
We evaluate five open-vocabulary mobile manipulation (OVMM) tasks representative of common household scenarios:
\textit{(1) Throw the can into the trash bin}
 \textit{(2) Move the pot near the red mug}
 \textit{(3) Put the mug on the shelf}
 \textit{(4) Open the cabinet}
\textit{(5) Open the dishwasher}

Tasks (1)–(3) are inspired by standard pick-and-place setups frequently explored in prior OVMM work~\cite{yenamandra2023homerobot, huang2023ok, qiu2024open, zhi2025closedloopopenvocabularymobilemanipulation}. However, we argue that OVMM should extend beyond pick-and-place to capture a broader range of interactions relevant to real-world deployment. Consequently, tasks (4) and (5) incorporate articulation-based actions to open objects, thereby expanding the diversity of interactions. These tasks cover a diverse range of object types with varying spatial and directional constraints. We include: (i) simple objects with relaxed requirements (e.g., cans and bins) that can be approached from most directions; (ii) objects like mugs, pots, and shelves that require alignment with specific affordance regions such as handles or flat surfaces; and (iii) articulated objects such as cabinets and dishwashers that require correct approach angles for successful manipulation. 

Each task is executed 20 times with randomized object placements, orientations, or initial robot base positions to evaluate robustness. A fixed random seed is used to ensure reproducibility. 

\subsection{Baseline Methods}

We compare our approach against four baselines, including classical geometric planners, keypoint-guided methods, and VLM-based prompting strategies. Except for \emph{Object Center + A*/RRT*}, all baselines share our coarse navigation setup: using A* to approach within 1.5\,m of the object, facing it. 

\begin{itemize}
    \item \textbf{Object Center + A*/RRT*}: Classical path planners that select a base placement at a fixed distance from the center of the target object, based solely on collision avoidance and proximity.

   \item \textbf{Affordance Point + A*/RRT*}: A VLM selects an affordance point $\mathbf{g}$ using the same procedure as our method, and a classical planner (A*/RRT*) computes a base placement at a fixed distance from $\mathbf{g}$ based on collision-free feasibility.

    \item \textbf{Pivot (\(I\))}:  Based on PIVOT~\cite{nasiriany2024pivot}, this method uses a VLM to iteratively update the placement action space, using only the RGB image and prompt. 

    \item \textbf{Pivot (\(M_{\text{local}}^+\), \(I_{\text{aff}}\))}: A variant of the above method that uses the same multimodal inputs as our approach. The VLM selects base placements using both \(M_{\text{local}}^+\) and \(I_{\text{aff}}\).
\end{itemize}

\subsection{Comparison Results}

\begin{figure}[htbp]
    \centering
    \includegraphics[width=0.97\linewidth]{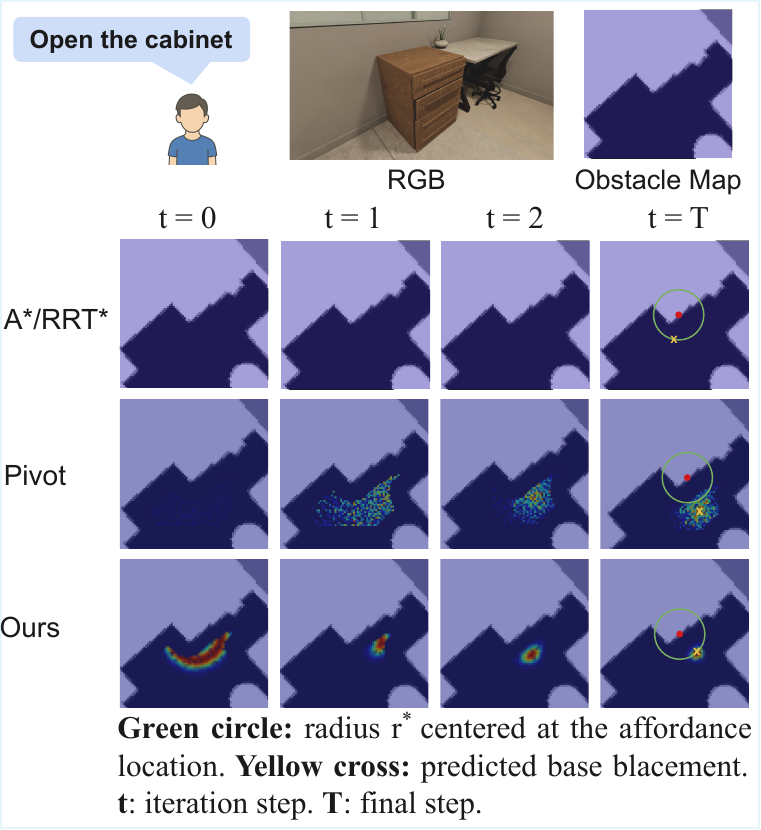}
    \caption{
Base placement distribution evolution for the task ``Open the cabinet.''
The A*/RRT* baseline (top row) selects a base placement at an oblique angle in front of the cabinet, which is not ideal.
The Pivot baseline (second row) selects a region in front of the cabinet but fails due to excessive distance from the target.
Our method (bottom row) converges to a base placement that is both feasible and semantically appropriate.
}
    \label{fig:heatmap}
\end{figure}

Table~\ref{tab:task_success} summarizes success rates across five semantic manipulation tasks. Our method achieves the highest overall success rate of 85\%, significantly outperforming all baselines. It excels on direction-sensitive tasks such as \textit{Open the cabinet}, while remaining robust on less constrained ones like \textit{Throw the can into the trash bin}. Figure~\ref{fig:heatmap} presents qualitative comparisons of base placements generated by different baselines and our method for the \textit{Open the cabinet} task, illustrating the importance of both semantic understanding and geometric feasibility.

\noindent\textbf{Object Center + A*/RRT*}\quad Classical geometric planners such as A* and RRT* rely solely on spatial proximity. They perform well in tasks where simply reaching a reasonable distance from the target is sufficient, such as \textit{ throwing the can into the trash bin} but fail when specific approach angles or semantic understanding are required. Due to their lack of semantic reasoning, success rates drop to 47\%  and 50\%.

\noindent\textbf{Affordance Point + A*/RRT*}\quad The limited performance of prior methods may be due to their tendency to approach the object as a whole, rather than targeting the manipulable part. To address this, we incorporate VLM-guided affordance point $\mathbf{g}$ to guide base placement. This yields modest improvements, particularly on spatially structured tasks, with overall success rates reaching up to 61\%. However, these methods still lack the ability to reason about directional constraints and reachability. Even when $\mathbf{g}$ lies near a handle, approaching from the side often results in failure. Mislocalization to occluded or non-manipulable regions further reduces effectiveness.

\noindent\textbf{Pivot (\(I\)) and Pivot (\(M_{\text{local}}^+\), \(I_{\text{aff}}\))}\quad To further enhance semantic reasoning, we consider VLM-driven prompting strategies. Pivot (\(I\)), which relies solely on RGB input, lacks geometric knowledge and often proposes placements that are either at incorrect distances or physically inaccessible. As a result, its performance is limited to 26\%. The multimodal variant, Pivot (\(M_{\text{local}}^+\), \(I_{\text{aff}}\)), incorporates affordance RGB and obstacle maps, which partially address perceptual gaps. Yet it still struggles to propose consistently feasible placements, with only a 23\% success rate.

\noindent\textbf{Our method}\quad Our method addresses the limitations of egocentric perception by projecting affordance cues onto obstacle maps through a cross-modal representation. We then perform a coarse-to-fine optimization that balances semantic intent and geometric feasibility. This approach leads to robust base placements and achieves an overall success rate of 85\%, significantly outperforming all baselines.

\subsection{Alpha Comparison}
We compare four configurations of the coefficient $\alpha_t$ in Eq.\eqref{eq:w(x)}, including fixed values of $0$, $0.5$, and $1$, as well as a sigmoid schedule defined in Eq.\eqref{eq:sigmoid}, which gradually increases $\alpha_t$ during optimization. This coefficient controls the weighting between semantic and geometric factors, allowing us to observe how different emphases affect task performance. The results confirm that the coarse-to-fine strategy yields the best overall outcome.

As shown in Table~\ref{tab:alpha_ablation}, setting $\alpha = 0$ (semantics only) results in sampling concentrated around task-relevant regions but often yields physically infeasible candidates, such as those beyond the robot’s reachable workspace. When $\alpha = 0.5$ (balanced semantic and geometric weights), the introduction of geometric constraints effectively filters out physically invalid candidates, leading to a substantial improvement in success rates. However, maintaining a fixed trade-off between semantic and geometric weights can still result in suboptimal placements, particularly in scenarios where the two objectives conflict. With $\alpha = 1$ (geometry only), sampling concentrates on a ring around the affordance point $\mathbf{g}$, favoring geometrically feasible placements. However, these sampled candidates may lack semantically appropriate options or include only a few such candidates. As a result, the VLM is more likely to select semantically incorrect base placements, causing the optimization to converge prematurely to geometrically valid but semantically misaligned regions and ultimately resulting in task failure.

Prior experiments showed that fixed weighting strategies are often insufficient. Semantic-only configurations tend to propose placements that are not physically operable on the target object, while geometry-only configurations risk overlooking the task intent. Even a balanced weighting can lead to suboptimal results when semantic and geometric objectives conflict. To overcome these issues, our approach prioritizes semantic alignment in the early stages, guiding the search toward task-relevant regions, and gradually shifts toward geometric precision to ensure physical feasibility. 

\begin{figure}[htbp]
    \centering   
    \includegraphics[width=1\linewidth]{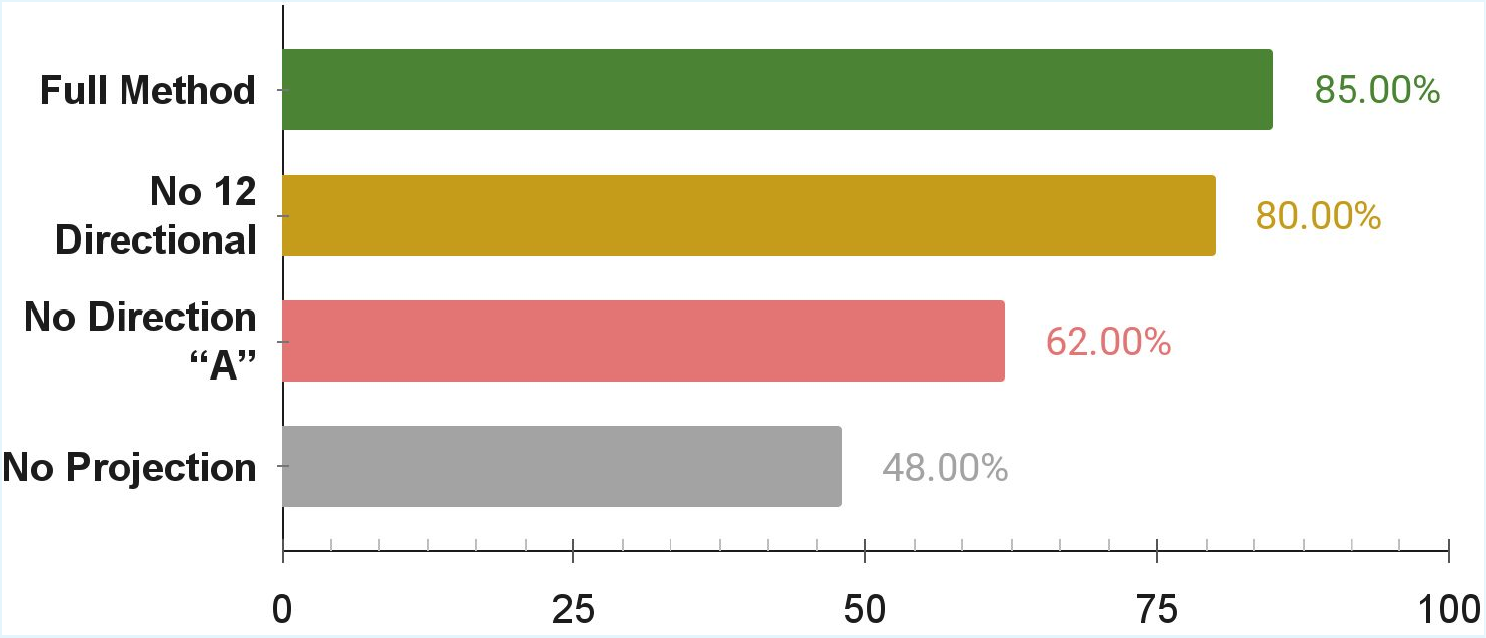}
\caption{
Projection Module Ablation.
The full method achieves 85\% success. Removing the 12 arrows causes a small drop (80\%), while removing the main arrow “A” leads to a larger drop (62\%). Without projection, performance drops most significantly to 48\%.
}

    \label{fig:Success Rates Of Affordance Guidance Ablation Variant}
\end{figure}

\subsection{Ablation on Affordance Guidance Projection}

We conducted an ablation study to evaluate the role of each component in our affordance-guidance projection module by comparing four system variants: (1) \textbf{Full Method}, which includes all components and projections; (2) \textbf{No 12 Arrows}, which retains ``A'' and $\mathcal{F}_t$, but removes the 12 auxiliary directional arrows; (3) \textbf{No Direction ``A''}, which removes the coarse affordance directional arrow ``A'', omitting the corresponding fan-shaped region $\mathcal{F}_t$; and (4) \textbf{No Projection}, which disables the entire projection mechanism and uses only RGB and raw obstacle maps. 

As shown in Figure~\ref{fig:Success Rates Of Affordance Guidance Ablation Variant}, the Full Method achieves the highest success rate at 85\%. Removing the 12 additional arrows while keeping ``A'' results in only a modest drop (to 80\%), indicating that while multi-directional cues provide some benefit, the single ``A'' arrow carries the main guidance signal. Eliminating the coarse arrow ``A'' reduces performance further (to 62\%), showing its critical role in base placement guidance. Finally, removing the projection leads to the largest performance drop (from 85\% to 48\%), confirming the importance of spatially grounding semantic cues.

Although current VLMs demonstrate strong semantic reasoning capabilities over RGB images, these findings suggest that they have limited ability to automatically transform such semantic understanding into spatially grounded reasoning. This limitation underscores the need for our explicit projection mechanism, which enables VLMs to better associate semantic intent with spatial context and perform more effective reasoning.

\section{Conclusion}

We identify a key challenge in open-vocabulary mobile manipulation (OVMM), where task failures are often caused by poor base placement. To address this, we propose a cross-modal representation that integrates RGB images with obstacle maps, overcoming the limitations of single-view perception. Our coarse-to-fine planning strategy balances semantics and geometry, enabling the robot to determine base placements that are both task-relevant and physically valid. The proposed method achieves an 85\% success rate across five open-vocabulary tasks.

\noindent\textbf{Limitations and Future Work.} While generally effective, the predicted placements may exhibit limited geometric precision compared to geometry-based methods, particularly in tasks that require accurate distance estimation. To help the vision-language model better understand semantics and spatial context, we use affordance-guided projection to align the RGB input with the obstacle map. However, the system can still produce incorrect reasoning in some cases. Nevertheless, we believe this limitation will diminish as VLMs continue to improve. Furthermore, although our method can identify base placements that are both semantically appropriate and geometrically feasible, the manipulator arm may still collide with obstacles during execution in more confined or cluttered environments. In future work, we aim to incorporate arm trajectory feasibility into the optimization process to ensure fully executable end-to-end motion plans.

\section*{Acknowledgements}
This work was supported in part by National Science and Technology Council, Taiwan, under Grant NSTC 113-2634-F-002-007. We are grateful to the National Center for High-performance Computing.
\bibliography{refs}

\ifshowappendix
\include{appendix}

\fi

\end{document}

%% file: appendix.tex
\appendix

\section{Map Representation}
\label{appendix: map representation}

The robot operates in a 2D environment represented by a global top-down occupancy grid map of size $200 \times 200$, with a spatial resolution of $0.05\,\text{m}$ per cell. This global map $M_{\text{global}}$ encodes both obstacle regions and object locations, all expressed in a world-aligned coordinate frame.

At runtime, the robot extracts a local egocentric map $M_{\text{local}}$ centered at its current base position. This local map is rotated to align with the robot’s forward-facing direction, providing a consistent spatial reference for downstream planning and decision making.

\section{Affordance Guidance Projection}
\label{appendix: Affordance Projection Pipeline}

\begin{figure*}[htbp]
    \centering
  
     \subfigure[Affordance Direction Selection]{
        \includegraphics[width=0.48\linewidth]{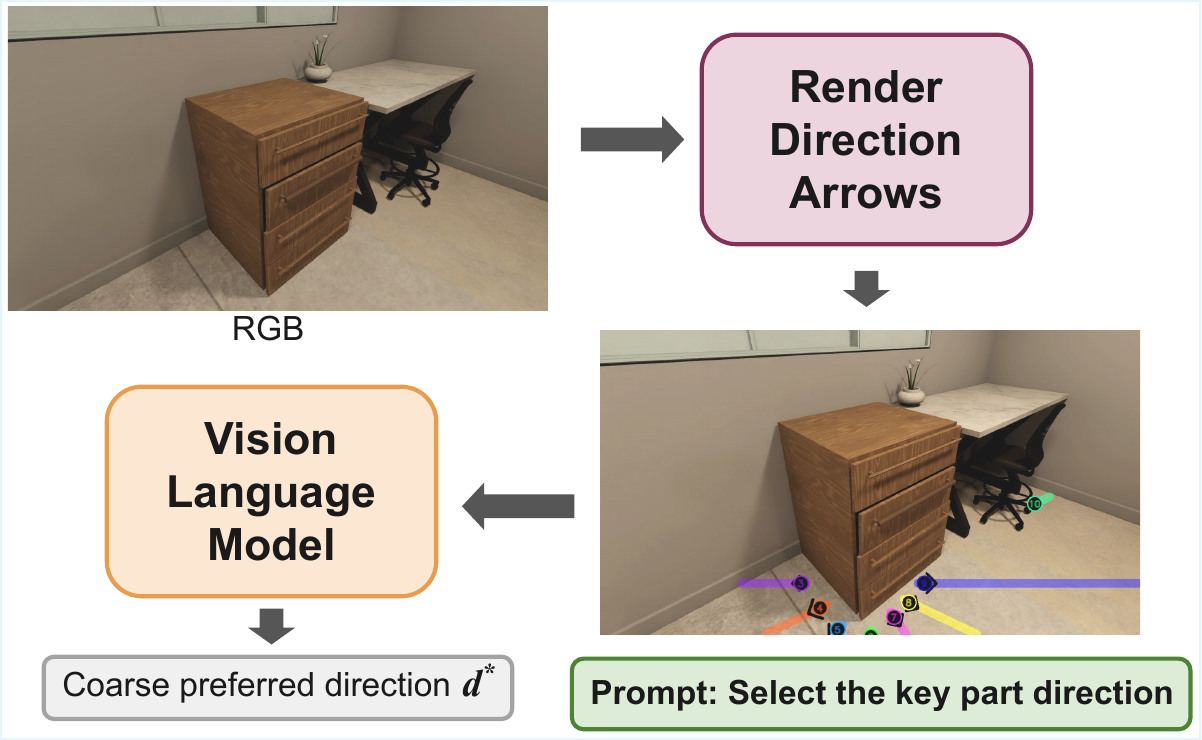}
        \label{fig:select direction}
    }
    \hfill
      \subfigure[Affordance Point Selection]{
        \includegraphics[width=0.48\linewidth]{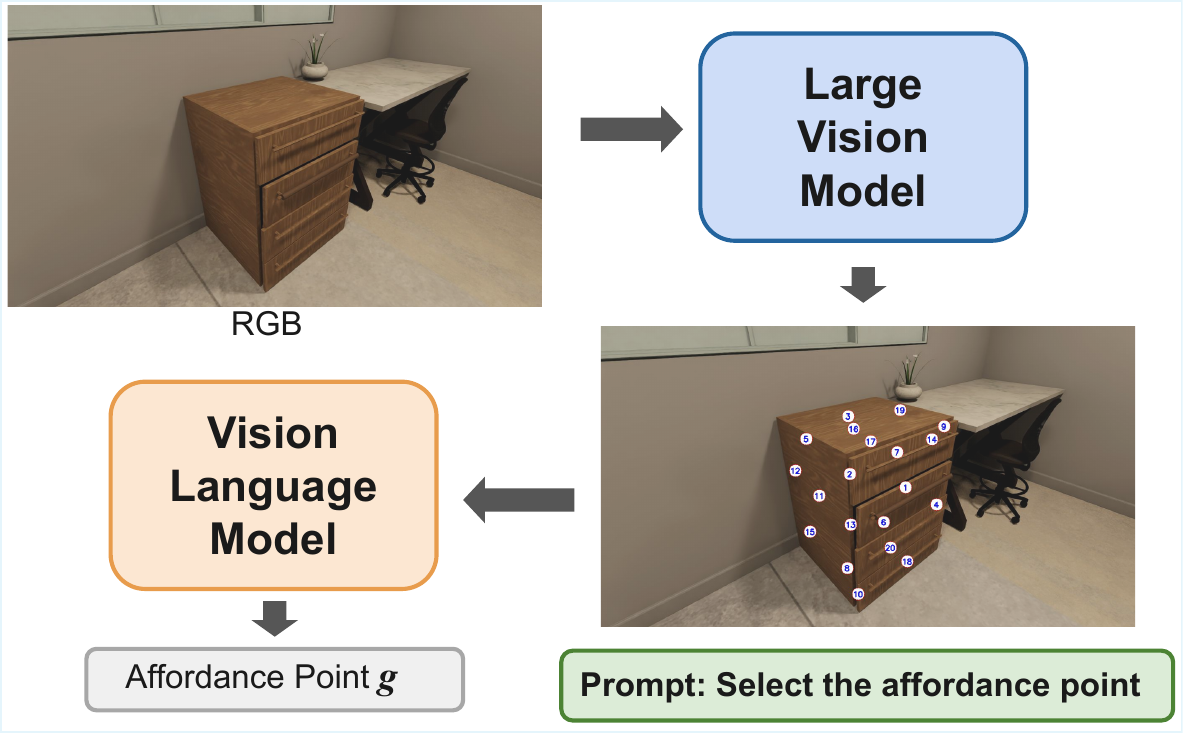}
        \label{fig:select affordance point}
    }
   
    \caption{
\textbf{Affordance guidance modules.} 
(a) Coarse affordance direction selection for the Affordance Guidance Projection. 
(b) Affordance point selection for the Affordance-Driven Coarse-to-Fine Optimization.
}
    
\end{figure*}

\begin{figure*}[htbp]
    
    \centering
    \includegraphics[width=0.95\linewidth]{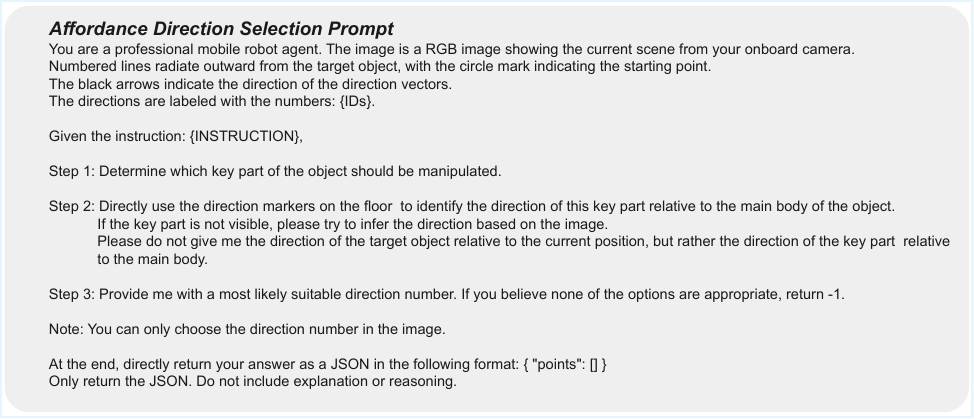}
    \caption{
      \textbf{Text prompt provided to VLM for selecting the semantically preferred affordance direction.} The vision-language model (VLM) is given an image of the target object annotated with 12 direction arrows uniformly placed around the object center and labeled with IDs. An instruction is also provided. The VLM is first asked to identify the key part of the object to be manipulated based on the instruction, and then to determine the direction of that part relative to the object’s main body using the direction markers.
    }
   \label{fig:direaction selection prompt}
\end{figure*}

We project language-conditioned affordance cues from image space into the map space through the following four steps:

\begin{enumerate}
    \item \textbf{Object Segmentation and Projection:}  
    Given the target object \( t \) and sub-instruction \( \tilde{\ell} \), we apply GroundedSAM to segment the object from the RGB image. The output mask is used to index the corresponding pixels in the aligned depth image \( D \). Using the camera intrinsics, each masked pixel is back-projected into a 3D point in the camera frame, then transformed into the robot base frame using the known extrinsics. The resulting 3D point cloud is projected onto the obstacle map $M_{\text{local}}$ by discretizing each point’s \((x, y)\) coordinates into map grid cells. The union of these cells defines the object footprint region \( \mathcal{R}_t \subseteq M_{\text{local}} \). We compute the centroid of this region as:
    \[
    \mathbf{c}_t = \frac{1}{|\mathcal{R}_t|} \sum_{(u,v) \in \mathcal{R}_t} (u, v),
    \]
    which serves as the origin for candidate direction rendering.

    \item \textbf{Robot Position Projection:}  
    The robot’s current base position \( \mathbf{p}_{\text{robot}} \in \mathbb{R}^2 \) is recorded in the same world-aligned coordinate frame and projected into map space using the same discretization as the object projection. This provides a reference point for spatial grounding in $M_{\text{local}}$ and helps VLMs interpret relative placement between the robot and the object.

   \item \textbf{Affordance Direction Selection:}  
We define a fixed set of 12 direction vectors \( \{ \mathbf{d}_i \}_{i=1}^{12} \), uniformly spaced every \(30^\circ\) around the centroid \( \mathbf{c}_t \). For each \( \mathbf{d}_i \), we draw an arrow from \( \mathbf{c}_t \) outward into the free space \( X_{\text{free}} \), using a predefined arrow length (e.g., 3.0\,m) and angle. These arrows lie on the floor surface and are rendered in the RGB image using known camera intrinsics and extrinsics, ensuring that all 12 directions are visible and labeled with their corresponding indices (1--12). The rendered RGB image—with all arrows and direction labels—is combined with the prompt containing instruction \( \tilde{\ell} \) and submitted to a vision-language model (VLM). The VLM is queried three times, returning a direction index \( i^{(k)} \in \{1,\dots,12\} \cup \{-1\} \), where \( -1 \) denotes ``uncertain'' or ``none of the above.'' The final direction is selected via majority vote:
\[
i^* = \operatorname{mode}(\{ i^{(1)}, i^{(2)}, i^{(3)} \}).
\]
If no majority exists or if \( i^* = -1 \), the selection is discarded and the trial is skipped. Otherwise, direction \( \mathbf{d}_{i^*} \) is selected as the task-preferred approach direction.  
The overall process is illustrated in Figure~\ref{fig:select direction}.  
The language prompt provided to the VLM is shown in Figure~\ref{fig:direaction selection prompt}.

\item \textbf{Affordance Region Projection:}  
Based on the selected direction \( \mathbf{d}_{i^*} \), we construct a fan-shaped affordance region \( \mathcal{F}_t \subseteq M \), centered at \( \mathbf{c}_t \), spanning \( \pm 60^\circ \) around \( \mathbf{d}_{i^*} \):
\[
\mathcal{F}_t = \left\{ \mathbf{x} \in M \;\middle|\; \angle(\mathbf{x} - \mathbf{c}_t, \mathbf{d}_{i^*}) \leq 60^\circ \right\}.
\]
This region captures directions that are semantically favorable for base approach. To ensure consistent reasoning across modalities, all 12 candidate direction arrows \( \{ \mathbf{d}_i \} \) are rendered in both the RGB image and the top-down obstacle map. Each arrow is assigned a fixed, unique color (e.g., direction 1 is always red, direction 2 always blue, etc.), and the selected direction \( \mathbf{d}_{i^*} \) is highlighted with the label `A'' in both views. This color-consistent dual rendering enables the VLM to associate visual cues in the RGB image with spatial semantics in the obstacle map.

\end{enumerate}

\section{Affordance Point Selection}
\label{Appendix: Affordance Point}

To determine the affordance point $\mathbf{g}$, we adopt a keypoint selection process inspired by \cite{huang2024rekep}. This process generates a set of visual keypoint proposals guided by object-level semantics:

\begin{enumerate}
    \item Extract patch-level features from the RGB image $I \in \mathbb{R}^{H \times W \times 3}$ using a vision backbone (e.g., DINOv2), yielding $\mathbf{F}_{\text{patch}} \in \mathbb{R}^{h' \times w' \times d}$. These are upsampled to the original resolution as $\mathbf{F}_{\text{interp}} \in \mathbb{R}^{H \times W \times d}$.
    
    \item Segment the target object $t$ using GroundedSAM to obtain a binary mask $S_t \in \{0, 1\}^{H \times W}$, and extract object features $\mathbf{F}_{\text{target}} = \mathbf{F}_{\text{interp}}[S_t]$.

    \item Apply $k$-means clustering ($k=20$) with cosine similarity to $\mathbf{F}_{\text{target}}$ to generate candidate visual keypoints. Each cluster centroid is back-projected to 3D, and candidates with pairwise distances below 0.08\,m are pruned to ensure spatial diversity.

    \item Render each keypoint on the RGB image and annotate it. A vision-language model (VLM) selects the most semantically relevant one as the affordance point $\mathbf{g}$ using a prompt containing sub-instruction $\tilde{\ell}$ and the annotated RGB image.

\end{enumerate}

\begin{figure*}[htbp]
    \centering
    \includegraphics[width=0.95\linewidth]{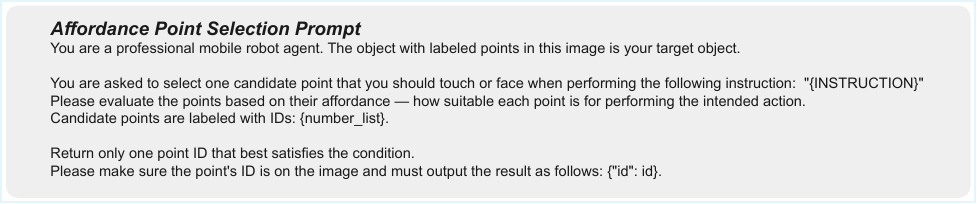}
    \caption{
    \textbf{Text prompt provided to VLM for selecting the affordance point.} 
    We present the vision-language model (VLM) with an image of the target object annotated with numbered candidate affordance points, along with an instruction. The VLM is asked to select the most appropriate point for the robot to touch or face in order to perform the given task.
    }

    \label{fig:affordance point selection prompt}
\end{figure*}

An overview of the affordance point selection process is illustrated in Figure~\ref{fig:select affordance point}. The language prompt provided to the VLM is shown in Figure~\ref{fig:affordance point selection prompt}.

\section{Baseline Methods}
\label{appendix:baselines}

We compare our approach against a range of baselines, including classical planners, keypoint-guided strategies, and VLM-based prompting. Except for A*/RRT*, all baselines adopt the same coarse navigation strategy as our method: using A* to move within 1.5\,m of the target object, oriented to face it. This standard initialization ensures consistent perception and spatial alignment prior to final base placement selection.

\begin{itemize}
    \item \textbf{Object Center + A*/RRT*}: Classical path planners that select a base placement from $\mathcal{X}_{\text{free}}$ by optimizing for a fixed preferred distance $r^* = 0.7$\,m to the center of the target object. These methods consider only geometric reachability and obstacle avoidance, without incorporating semantic understanding or task intent.

    \item \textbf{Affordance Point + A*/RRT*}: A semantic-aware variant in which a VLM selects a task-relevant affordance point $\mathbf{g}$ using the same procedure as our method. A*/RRT* then selects a collision-free base placement at the preferred distance $r^*$ from $\mathbf{g}$. 

  \item \textbf{Pivot ($I$)}: Based on PIVOT~\cite{nasiriany2024pivot}, this method uses a VLM to iteratively update the placement action space. Candidate base placements are rendered as annotated markers on the RGB image and input to the VLM as a language prompt containing the instruction. Unlike the original PIVOT, we constrain all candidates to lie within the collision-free set $\mathcal{X}_{\text{free}}$, ensuring physical feasibility in terms of base placement. 

\item \textbf{Pivot ($M_{\text{local}}^+$, $I_{\text{aff}}$)}: An extended version of Pivot that uses the same multimodal inputs as our method—including \textit{Obstacle Map+} ($M_{\text{local}}^+$), the \textit{Affordance RGB} ($I_{\text{aff}}$), and the language prompt. Like the RGB-only variant, all base candidates are restricted to $\mathcal{X}_{\text{free}}$ to avoid infeasible placements.
\end{itemize}

\section{Base placement distribution evolution}
\label{appendix:Base placement distribution evolution}

\begin{figure*}[htbp]
    
    \centering
    \includegraphics[width=0.9\linewidth]{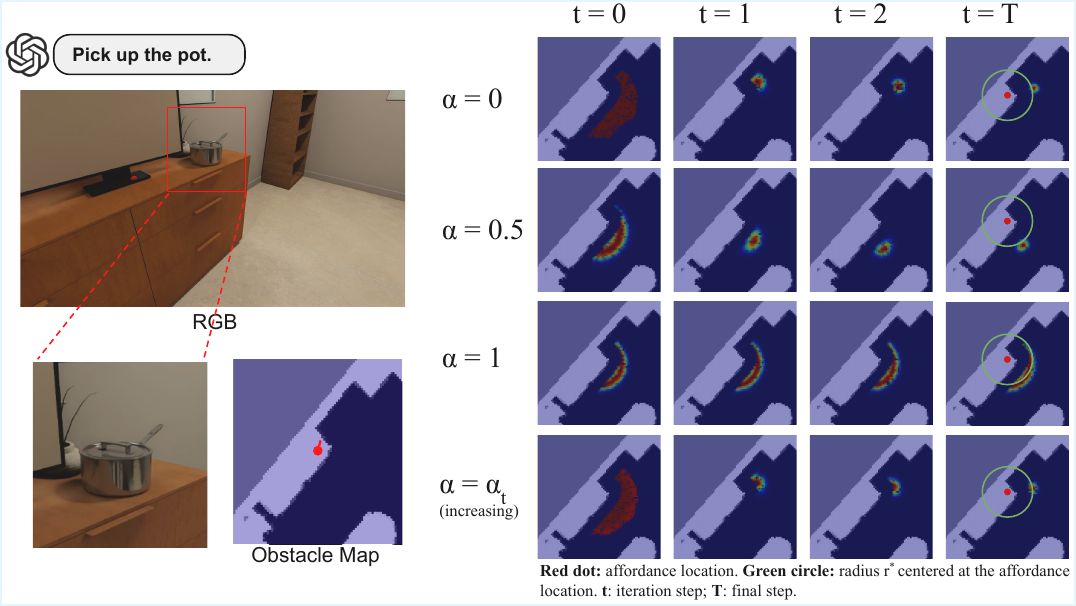}
    \caption{
        \textbf{Effect of the weighting coefficient \(\alpha_t\) on base placement optimization for the task ``Pick up the pot.''} 
        The heatmaps show sampled candidate distributions across iterations \(t = 0 \rightarrow T\) under four settings:
        \textbf{\(\alpha = 0\)} (semantic-only),
        \textbf{\(\alpha = 0.5\)} (fixed balance),
        \textbf{\(\alpha = 1\)} (geometry-only), and
        \textbf{\(\alpha = \alpha_t\)} (our coarse-to-fine schedule). Our method yields more precise and affordance-aligned placements.
    }
   \label{fig:alpha comparisom}
\end{figure*}

As shown in Fig.~\ref{fig:alpha comparisom}, we observe that in the $\alpha = 0$ setting (semantic-only), the candidate distribution initially concentrates around the handle side. Although this configuration performs reasonably well in this specific task, the final placement remains slightly farther from the preferred radius $r^*$, indicating limited geometric precision. Under $\alpha = 0.5$, the optimization converges to a placement that is neither semantically meaningful nor spatially ideal, suggesting that a static balance between semantics and geometry may dilute the strengths of both. In the $\alpha = 1$ case (geometry-only), the distribution remains nearly unchanged across iterations due to the lack of semantic updates. While the candidates are well-aligned with the preferred radius $r^*$ relative to the affordance point $\mathbf{g}$. Most candidates tend to cluster on the side of the pot, often resulting in placements that are geometrically feasible but semantically misaligned. As a result, the VLM frequently selects candidates on the incorrect side—away from the handle—making it difficult for the robot to perform the intended manipulation.

By contrast, our dynamic $\alpha_t$ strategy successfully guides the optimization to a placement that balances semantic relevance and geometric accuracy, achieving a final candidate location that is both semantically meaningful and spatially appropriate.

\begin{algorithm}
\caption{Coarse-to-Fine Optimization}
\label{alg:affordance_sampling}
\KwIn{Affordance point $\mathbf{g}$, Affordance RGB $I_{\text{aff}}$, map $M_{\text{local}}^+$, instruction $\tilde{\ell}$}
\KwOut{Final base placement $x^*$}

\vspace{0.5em}

\textbf{Initialize:} $\mu_0 \leftarrow \emptyset$ \;

\For{$t = 1$ \KwTo $T$}{

    Compute $\alpha_t$ using Eq.~\eqref{eq:sigmoid} \;

    \textbf{Sampling and Scoring} \\
    \Indp Sample $N$ candidates $\{x_i\}$ from $\mathcal{N}(\mathbf{g}, \sigma^2 I)$ \;
    Reject $x_i$ if $\|x_i - \mathbf{g}\| > r_{\text{max}}$ or $x_i \notin \mathcal{X}_{\text{free}}$ \;
    Compute $w(x_i)$ using Eq.~\eqref{eq:w(x)} \;
    Normalize using Eq.~\eqref{eq:prob} to get $p(x_i)$ \;
    Sample $N_{\text{sample}}$ candidates $\{x_i'\}$ from $p(x_i)$ \;
    \Indm

    \textbf{Semantic Ranking:} \\
    \Indp Overlay $\{x_i'\}$ on $M_{\text{local}}^+$ with indices \;
    Query VLM with $(M_{\text{local}}^+, I_{\text{aff}}, \tilde{\ell})$ \;

    \eIf{$t < T$}{
        Receive top-$k$ ranked candidates $\{x^{(i)}\}_{i=1}^k$ \;
        
        \textbf{Update:} \\
        \Indp Update semantic center: $\mu_t = \frac{1}{k} \sum_{i=1}^k x^{(i)}$ \;
        Decay $\sigma_s$ \;
        \Indm
    }{
        Receive top-5 ranked candidates $\{x^{(i)}\}_{i=1}^5$ \;
        
        \textbf{Final selection:} \\
        \Indp Remove 2 furthest from their mean \;
        Compute $x^* = \frac{1}{3} \sum_{i=1}^{3} x^{(i)}$ \;
        \textbf{Return} $x^*$ \;
        \Indm
    }
    \Indm
}
\end{algorithm}

\begin{table}[!htbp]
\centering
\caption{The hyperparameters used in base selection.}
\label{tab:hyperparams}
\begin{tabular}{ll}
\toprule
\textbf{Parameter} & \textbf{Value / Description} \\
\midrule
$N$ & 1000 (initial candidate samples per iteration) \\
$N_{\text{sample}}$ & 20 (resampled candidates per iteration) \\
$T$ & 4 (number of iterations) \\
$r_{\text{max}}$ & 1.2\,m (truncation radius) \\
$r^*$ & 0.7\,m (preferred geometric distance) \\
$\sigma_{\text{sample}}$ & 1.0\,m (initial sampling std. deviation) \\
$\sigma_g$ & 0.1\,m (geometric window std. deviation) \\
$\sigma_s$ & exponential decay: $0.8^t \cdot 0.2$ \\
$\alpha_{\text{max}}$ & 0.6 (maximum geometric weight) \\
$\gamma$ & 2.0 (sigmoid steepness for $\alpha_t$ schedule) \\
$\delta$ & 0.05\,m (margin in $\Phi(d; \mu, \sigma)$) \\
$K$ & 3 (top-$k$ points returned by VLM) \\
\bottomrule
\end{tabular}
\end{table}

\begin{figure}[htbp]
    \centering
    \includegraphics[width=1\linewidth]{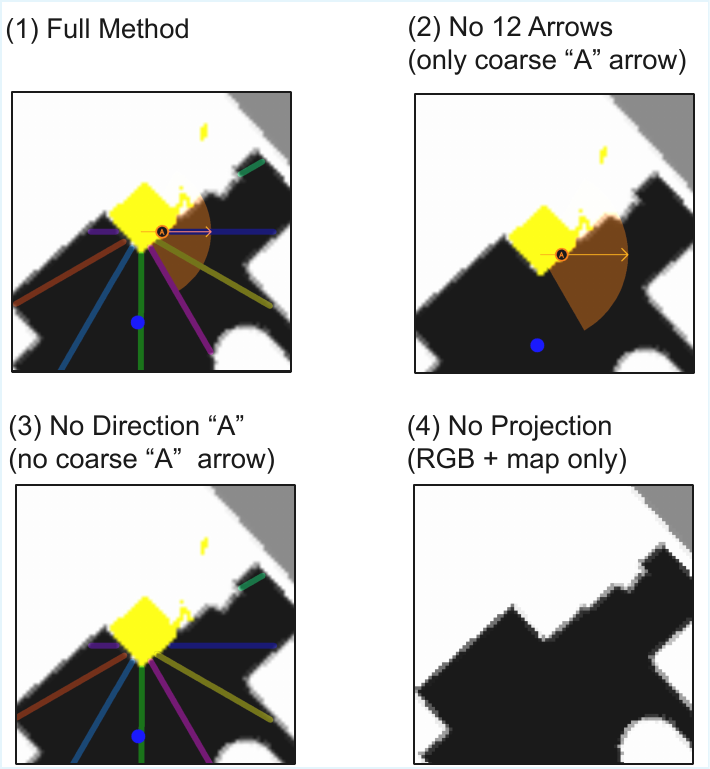}
    \caption{\textbf{Illustration of different variants of affordance guidance for the task ``Open the cabinet.''} (1) Full Method: includes all components; (2) No 12 Arrows: removes auxiliary arrows; (3) No Direction ``A'': removes the main directional cue; (4) No Projection: disables the projection mechanism.}

    \label{fig:Affordance Guidance Projection}
\end{figure}

\begin{figure}[htbp]
    \centering
  
     \subfigure[Affordance RGB]{
        \includegraphics[width=0.5\linewidth]{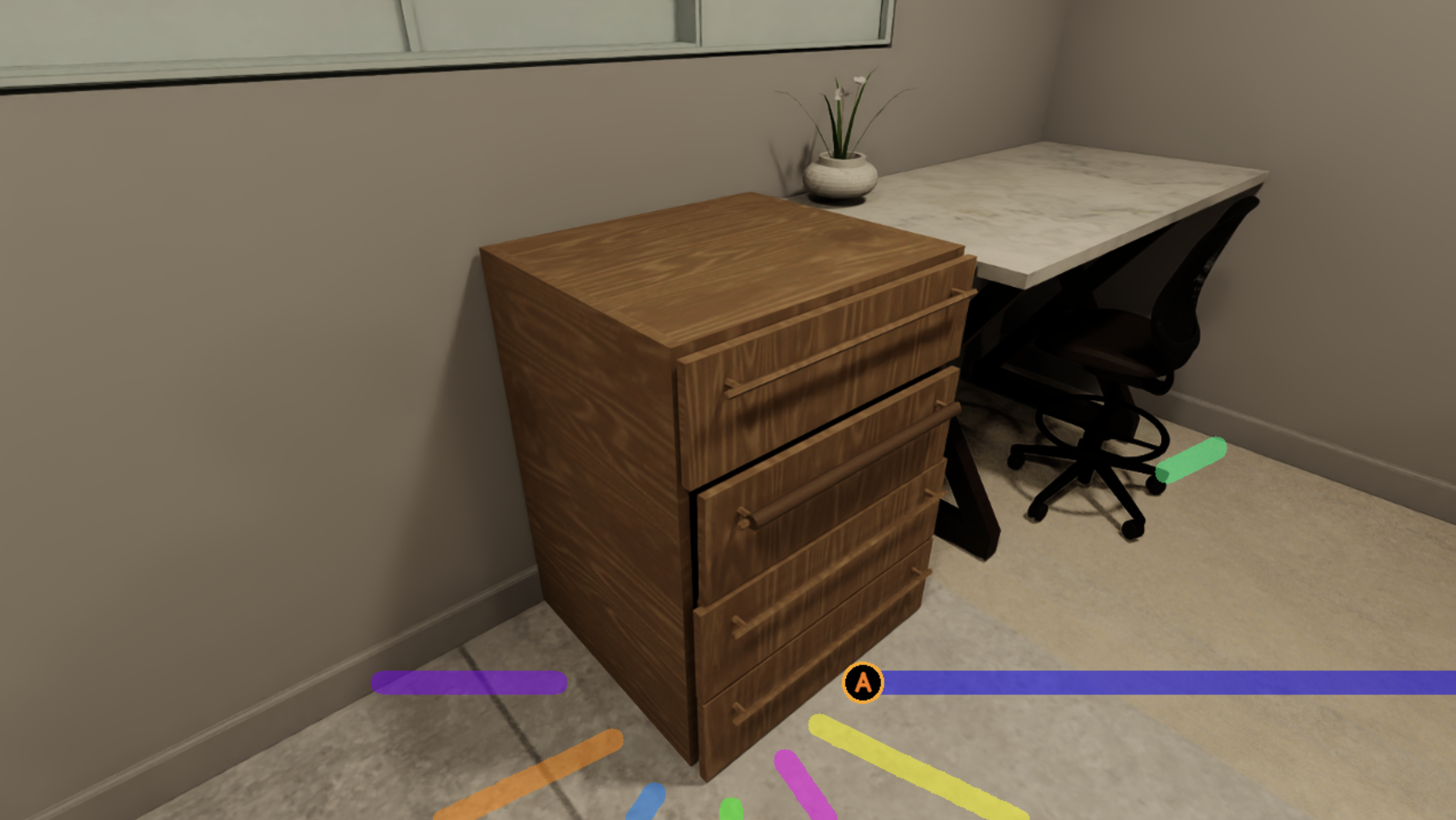}
        \label{fig:affordance_rgb}
    }
    \hfill
      \subfigure[Obstacle Map+]{
        \includegraphics[width=0.35\linewidth]{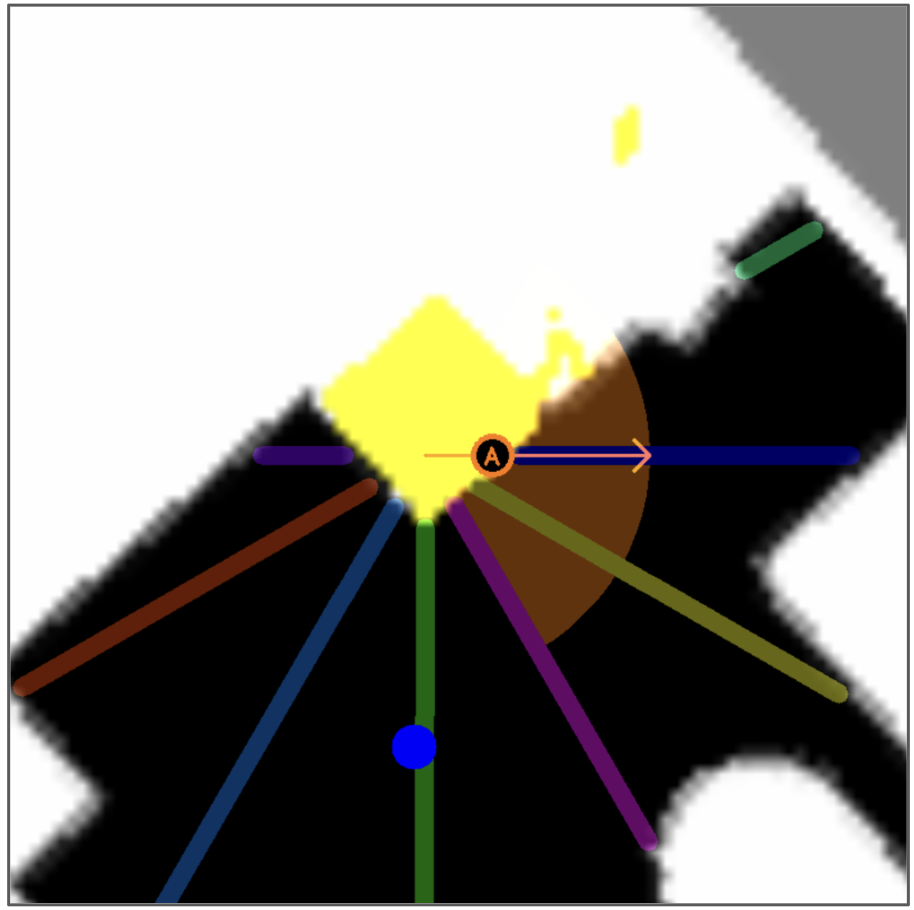}
        \label{fig:obstacle_map+}
    }
   
    \caption{
The two representations constructed by the Affordance Guidance Projection.
}
    
\end{figure}

\begin{figure}[htbp]
    \centering
  
     \subfigure[]{
        \includegraphics[width=0.3\linewidth]{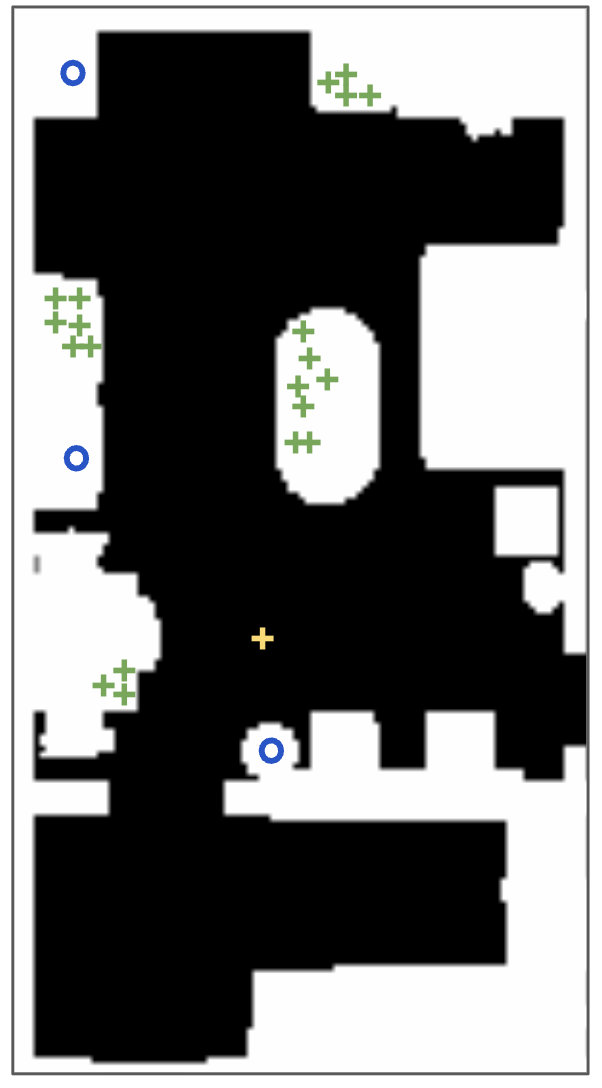}
        \label{fig:map1}
    }
    \hfill
      \subfigure[]{
        \includegraphics[width=0.3\linewidth]{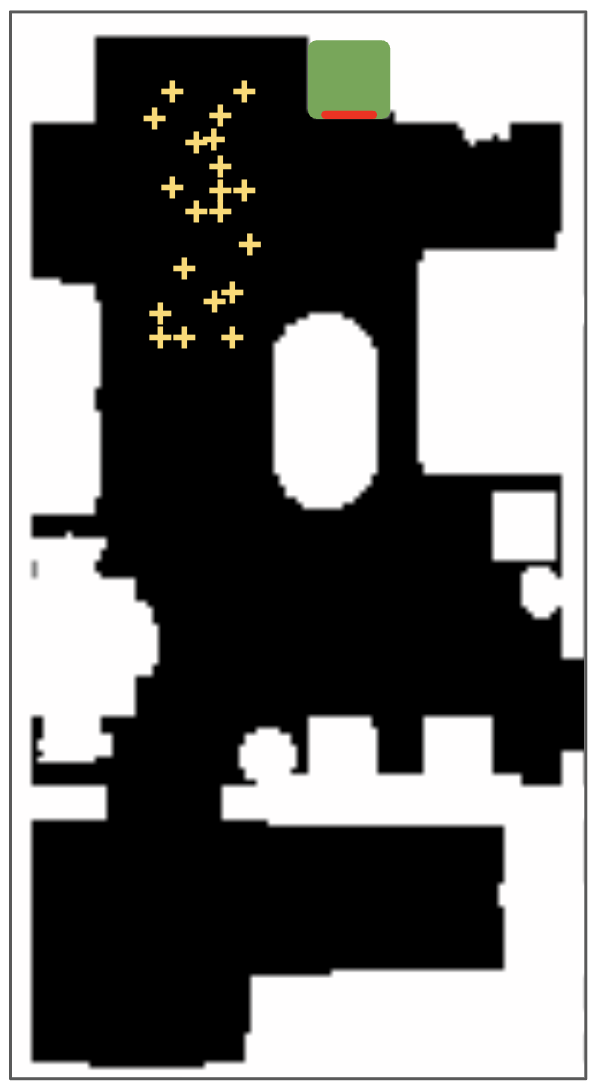}
        \label{fig:map2}
    }
    \hfill
      \subfigure[]{
        \includegraphics[width=0.3\linewidth]{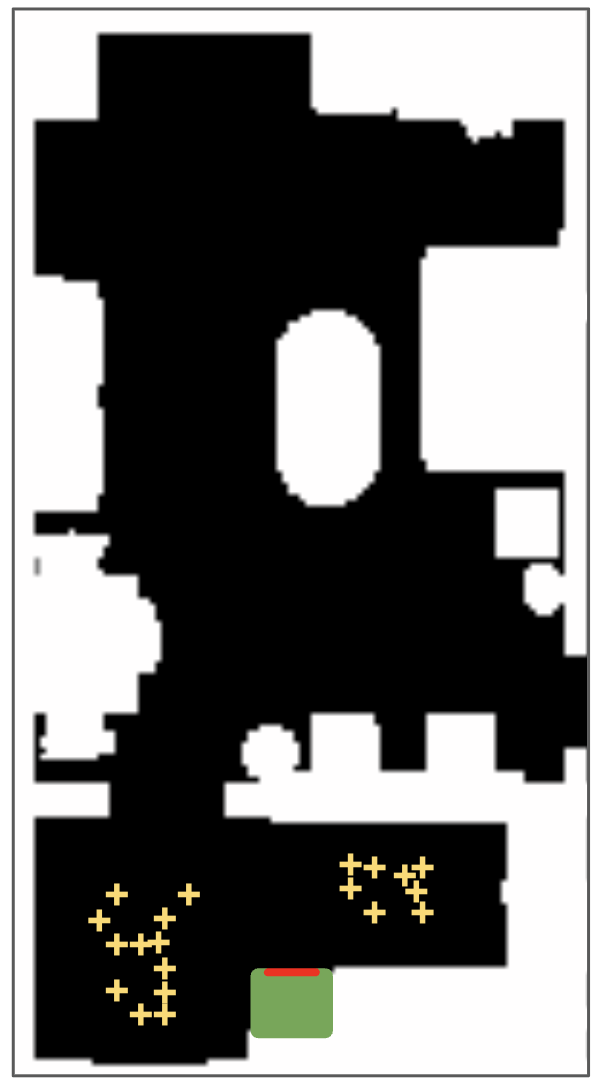}
        \label{fig:map3}
    }
   
  \caption{
\textbf{Initial Environment Setup of Tasks.}  
The figure illustrates the initial environment configurations.  
Yellow crosses denote the initial base distribution of the robot, and green crosses or blocks indicate the manipulation targets.
(a) Pick-and-place tasks: throw the can into the trash bin, move the pot near the red mug, and put the mug on the shelf. Blue circles indicate the placement goals (e.g., shelf, near the red mug, trash bin).  
(b) Open the cabinet.  
(c) Open the dishwasher.
}
\label{fig:Initial Environment Setup}
    
\end{figure}

\begin{figure*}[htbp]
    \centering
    \includegraphics[width=0.95\linewidth]{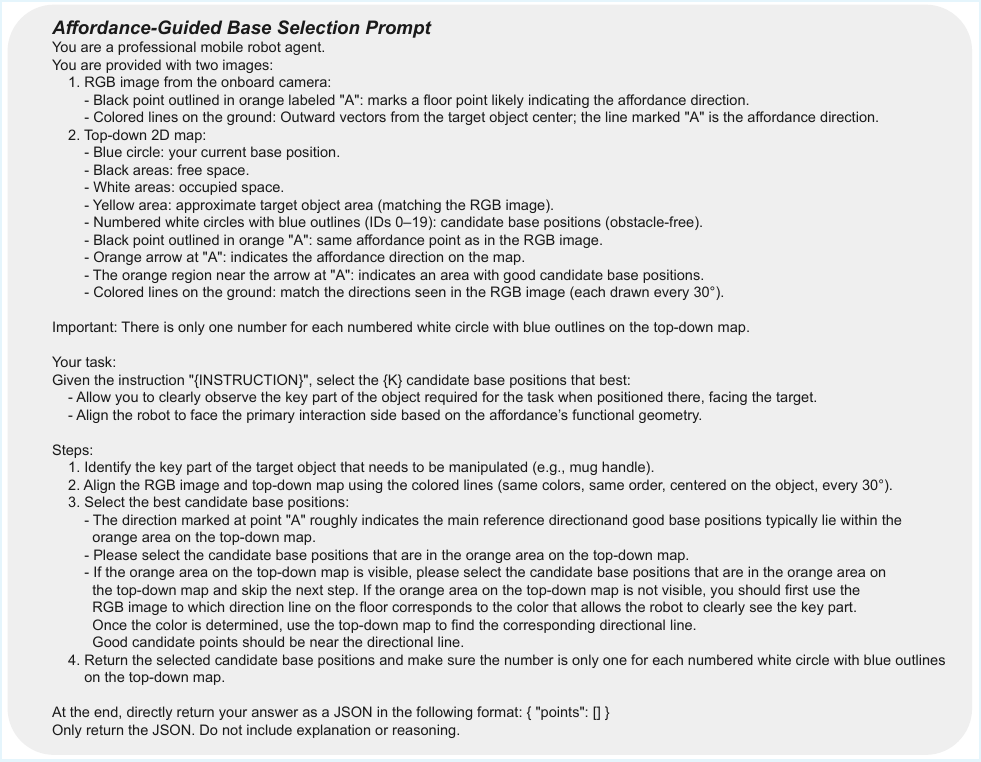}
    \caption{
     \textbf{Text prompt provided to the VLM for selecting an appropriate base placement.}
     The vision-language model (VLM) is provided with two inputs: (1) Affordance RGB, and (2) obstacle map+ showing candidate base placements (IDs 0–19). The prompt first describes the key visual features present in both images, including the affordance direction, spatial markers, and object location, to help the model align the Affordance RGB with the obstacle map+. The VLM is then instructed to select \textit{K} candidate base placements that allow clear observation and proper alignment with the manipulable part of the object. The prompt further guides the selection by referring to cues such as the orange reference region and color-matched directional vectors shared across both views.
    }
    \label{fig:base selection prompt}
\end{figure*}



%% file: refs.bib
@inproceedings{singh2023clipfields,
  title={CLIP-Fields: Weakly supervised semantic fields for robotic manipulation},
  author={Singh, Ajay M and Yang, Junhyuk and Chen, Anqi and Wu, Jiajun and Finn, Chelsea},
  booktitle={Conference on Robot Learning (CoRL)},
  year={2023}
}

@inproceedings{huang2023ok,
  title={OK-Robot: Zero-shot object navigation using multimodal world models},
  author={Huang, Jialin and Liang, Jack and Shi, Bowen and Yang, Yilun and Liu, Yilun and Driess, Danny and Toussaint, Marc and Fu, Kuan and Lin, Ke and Liu, Ziang and others},
  booktitle={Conference on Robot Learning (CoRL)},
  year={2023}
}

@inproceedings{huang2023visual,
  title={VLMaps: Grounding large language models with spatial maps for navigation},
  author={Huang, Jialin and Yang, Yilun and Driess, Danny and others},
  booktitle={Conference on Robot Learning (CoRL)},
  year={2023}
}

@article{nasiriany2024pivot,
  title={Pivot: Iterative visual prompting elicits actionable knowledge for vlms},
  author={Nasiriany, Soroush and Xia, Fei and Yu, Wenhao and Xiao, Ted and Liang, Jacky and Dasgupta, Ishita and Xie, Annie and Driess, Danny and Wahid, Ayzaan and Xu, Zhuo and others},
  journal={arXiv preprint arXiv:2402.07872},
  year={2024}
}

@article{qiu2024open,
  title={Open-vocabulary mobile manipulation in unseen dynamic environments with 3d semantic maps},
  author={Qiu, Dicong and Ma, Wenzong and Pan, Zhenfu and Xiong, Hui and Liang, Junwei},
  journal={arXiv preprint arXiv:2406.18115},
  year={2024}
}

@inproceedings{bolte2023usa,
  title={Usa-net: Unified semantic and affordance representations for robot memory},
  author={Bolte, Ben and Wang, Austin and Yang, Jimmy and Mukadam, Mustafa and Kalakrishnan, Mrinal and Paxton, Chris},
  booktitle={2023 IEEE/RSJ International Conference on Intelligent Robots and Systems (IROS)},
  pages={1--8},
  year={2023},
  organization={IEEE}
}

@article{huang2024rekep,
  title={Rekep: Spatio-temporal reasoning of relational keypoint constraints for robotic manipulation},
  author={Huang, Wenlong and Wang, Chen and Li, Yunzhu and Zhang, Ruohan and Fei-Fei, Li},
  journal={arXiv preprint arXiv:2409.01652},
  year={2024}
}

@article{hart1968formal,
  title={A formal basis for the heuristic determination of minimum cost paths},
  author={Hart, Peter E and Nilsson, Nils J and Raphael, Bertram},
  journal={IEEE transactions on Systems Science and Cybernetics},
  volume={4},
  number={2},
  pages={100--107},
  year={1968},
  publisher={IEEE}
}

@article{karaman2011sampling,
  title={Sampling-based algorithms for optimal motion planning},
  author={Karaman, Sertac and Frazzoli, Emilio},
  journal={The international journal of robotics research},
  volume={30},
  number={7},
  pages={846--894},
  year={2011},
  publisher={Sage Publications Sage UK: London, England}
}

@article{yenamandra2023homerobot,
  title={Homerobot: Open-vocabulary mobile manipulation},
  author={Yenamandra, Sriram and Ramachandran, Arun and Yadav, Karmesh and Wang, Austin and Khanna, Mukul and Gervet, Theophile and Yang, Tsung-Yen and Jain, Vidhi and Clegg, Alexander William and Turner, John and others},
  journal={arXiv preprint arXiv:2306.11565},
  year={2023}
}

@misc{chang2023goatthing,
      title={GOAT: GO to Any Thing}, 
      author={Matthew Chang and Theophile Gervet and Mukul Khanna and Sriram Yenamandra and Dhruv Shah and So Yeon Min and Kavit Shah and Chris Paxton and Saurabh Gupta and Dhruv Batra and Roozbeh Mottaghi and Jitendra Malik and Devendra Singh Chaplot},
      year={2023},
      eprint={2311.06430},
      archivePrefix={arXiv},
      primaryClass={cs.RO},
      url={https://arxiv.org/abs/2311.06430}, 
}

@misc{yang2023setofmarkpromptingunleashesextraordinary,
      title={Set-of-Mark Prompting Unleashes Extraordinary Visual Grounding in GPT-4V}, 
      author={Jianwei Yang and Hao Zhang and Feng Li and Xueyan Zou and Chunyuan Li and Jianfeng Gao},
      year={2023},
      eprint={2310.11441},
      archivePrefix={arXiv},
      primaryClass={cs.CV},
      url={https://arxiv.org/abs/2310.11441}, 
}

@misc{huang2024copageneralroboticmanipulation,
      title={CoPa: General Robotic Manipulation through Spatial Constraints of Parts with Foundation Models}, 
      author={Haoxu Huang and Fanqi Lin and Yingdong Hu and Shengjie Wang and Yang Gao},
      year={2024},
      eprint={2403.08248},
      archivePrefix={arXiv},
      primaryClass={cs.RO},
      url={https://arxiv.org/abs/2403.08248}, 
}

@misc{sathyamoorthy2024convoicontextawarenavigationusing,
      title={CoNVOI: Context-aware Navigation using Vision Language Models in Outdoor and Indoor Environments}, 
      author={Adarsh Jagan Sathyamoorthy and Kasun Weerakoon and Mohamed Elnoor and Anuj Zore and Brian Ichter and Fei Xia and Jie Tan and Wenhao Yu and Dinesh Manocha},
      year={2024},
      eprint={2403.15637},
      archivePrefix={arXiv},
      primaryClass={cs.RO},
      url={https://arxiv.org/abs/2403.15637}, 
}

@misc{shao2024momaposefficientobjectkinematicawarebase,
      title={MoMa-Pos: An Efficient Object-Kinematic-Aware Base Placement Optimization Framework for Mobile Manipulation}, 
      author={Beichen Shao and Nieqing Cao and Yan Ding and Xingchen Wang and Fuqiang Gu and Chao Chen},
      year={2024},
      eprint={2403.19940},
      archivePrefix={arXiv},
      primaryClass={cs.RO},
      url={https://arxiv.org/abs/2403.19940}, 
}

@misc{zhang2025momakitchen100kbenchmarkaffordancegrounded,
      title={MoMa-Kitchen: A 100K+ Benchmark for Affordance-Grounded Last-Mile Navigation in Mobile Manipulation}, 
      author={Pingrui Zhang and Xianqiang Gao and Yuhan Wu and Kehui Liu and Dong Wang and Zhigang Wang and Bin Zhao and Yan Ding and Xuelong Li},
      year={2025},
      eprint={2503.11081},
      archivePrefix={arXiv},
      primaryClass={cs.RO},
      url={https://arxiv.org/abs/2503.11081}, 
}

@misc{zhu2024navi2gazeleveragingfoundationmodels,
      title={Navi2Gaze: Leveraging Foundation Models for Navigation and Target Gazing}, 
      author={Jun Zhu and Zihao Du and Haotian Xu and Fengbo Lan and Zilong Zheng and Bo Ma and Shengjie Wang and Tao Zhang},
      year={2024},
      eprint={2407.09053},
      archivePrefix={arXiv},
      primaryClass={cs.RO},
      url={https://arxiv.org/abs/2407.09053}, 
}

@misc{openai2023gpt4,
  author = {{OpenAI}},
  title = {GPT-4 Technical Report},
  year = {2023},
  howpublished = {\url{https://arxiv.org/abs/2303.08774}},
  note = {Accessed June 6, 2025}
}

@misc{liu2023visualinstructiontuning,
      title={Visual Instruction Tuning}, 
      author={Haotian Liu and Chunyuan Li and Qingyang Wu and Yong Jae Lee},
      year={2023},
      eprint={2304.08485},
      archivePrefix={arXiv},
      primaryClass={cs.CV},
      url={https://arxiv.org/abs/2304.08485}, 
}

@article{team2023gemini,
  title={Gemini: a family of highly capable multimodal models},
  author={Team, Gemini and Anil, Rohan and Borgeaud, Sebastian and Alayrac, Jean-Baptiste and Yu, Jiahui and Soricut, Radu and Schalkwyk, Johan and Dai, Andrew M and Hauth, Anja and Millican, Katie and others},
  journal={arXiv preprint arXiv:2312.11805},
  year={2023}
}

@misc{radford2021learningtransferablevisualmodels,
      title={Learning Transferable Visual Models From Natural Language Supervision}, 
      author={Alec Radford and Jong Wook Kim and Chris Hallacy and Aditya Ramesh and Gabriel Goh and Sandhini Agarwal and Girish Sastry and Amanda Askell and Pamela Mishkin and Jack Clark and Gretchen Krueger and Ilya Sutskever},
      year={2021},
      eprint={2103.00020},
      archivePrefix={arXiv},
      primaryClass={cs.CV},
      url={https://arxiv.org/abs/2103.00020}, 
}

@misc{kirillov2023segment,
      title={Segment Anything}, 
      author={Alexander Kirillov and Eric Mintun and Nikhila Ravi and Hanzi Mao and Chloe Rolland and Laura Gustafson and Tete Xiao and Spencer Whitehead and Alexander C. Berg and Wan-Yen Lo and Piotr Dollár and Ross Girshick},
      year={2023},
      eprint={2304.02643},
      archivePrefix={arXiv},
      primaryClass={cs.CV},
      url={https://arxiv.org/abs/2304.02643}, 
}

@misc{minderer2022simpleopenvocabularyobjectdetection,
      title={Simple Open-Vocabulary Object Detection with Vision Transformers}, 
      author={Matthias Minderer and Alexey Gritsenko and Austin Stone and Maxim Neumann and Dirk Weissenborn and Alexey Dosovitskiy and Aravindh Mahendran and Anurag Arnab and Mostafa Dehghani and Zhuoran Shen and Xiao Wang and Xiaohua Zhai and Thomas Kipf and Neil Houlsby},
      year={2022},
      eprint={2205.06230},
      archivePrefix={arXiv},
      primaryClass={cs.CV},
      url={https://arxiv.org/abs/2205.06230}, 
}

@misc{ren2024groundedsamassemblingopenworld,
      title={Grounded SAM: Assembling Open-World Models for Diverse Visual Tasks}, 
      author={Tianhe Ren and Shilong Liu and Ailing Zeng and Jing Lin and Kunchang Li and He Cao and Jiayu Chen and Xinyu Huang and Yukang Chen and Feng Yan and Zhaoyang Zeng and Hao Zhang and Feng Li and Jie Yang and Hongyang Li and Qing Jiang and Lei Zhang},
      year={2024},
      eprint={2401.14159},
      archivePrefix={arXiv},
      primaryClass={cs.CV},
      url={https://arxiv.org/abs/2401.14159}, 
}

@misc{oquab2024dinov2learningrobustvisual,
      title={DINOv2: Learning Robust Visual Features without Supervision}, 
      author={Maxime Oquab and Timothée Darcet and Théo Moutakanni and Huy Vo and Marc Szafraniec and Vasil Khalidov and Pierre Fernandez and Daniel Haziza and Francisco Massa and Alaaeldin El-Nouby and Mahmoud Assran and Nicolas Ballas and Wojciech Galuba and Russell Howes and Po-Yao Huang and Shang-Wen Li and Ishan Misra and Michael Rabbat and Vasu Sharma and Gabriel Synnaeve and Hu Xu and Hervé Jegou and Julien Mairal and Patrick Labatut and Armand Joulin and Piotr Bojanowski},
      year={2024},
      eprint={2304.07193},
      archivePrefix={arXiv},
      primaryClass={cs.CV},
      url={https://arxiv.org/abs/2304.07193}, 
}

@misc{lee2025affordanceguidedreinforcementlearningvisual,
      title={Affordance-Guided Reinforcement Learning via Visual Prompting}, 
      author={Olivia Y. Lee and Annie Xie and Kuan Fang and Karl Pertsch and Chelsea Finn},
      year={2025},
      eprint={2407.10341},
      archivePrefix={arXiv},
      primaryClass={cs.RO},
      url={https://arxiv.org/abs/2407.10341}, 
}

@misc{laina2025findanythingopenvocabularyobjectcentricmapping,
      title={FindAnything: Open-Vocabulary and Object-Centric Mapping for Robot Exploration in Any Environment}, 
      author={Sebastián Barbas Laina and Simon Boche and Sotiris Papatheodorou and Simon Schaefer and Jaehyung Jung and Stefan Leutenegger},
      year={2025},
      eprint={2504.08603},
      archivePrefix={arXiv},
      primaryClass={cs.RO},
      url={https://arxiv.org/abs/2504.08603}, 
}

@misc{zhi2025closedloopopenvocabularymobilemanipulation,
      title={Closed-Loop Open-Vocabulary Mobile Manipulation with GPT-4V}, 
      author={Peiyuan Zhi and Zhiyuan Zhang and Yu Zhao and Muzhi Han and Zeyu Zhang and Zhitian Li and Ziyuan Jiao and Baoxiong Jia and Siyuan Huang},
      year={2025},
      eprint={2404.10220},
      archivePrefix={arXiv},
      primaryClass={cs.RO},
      url={https://arxiv.org/abs/2404.10220}, 
}

@misc{tan2025languageconditionedopenvocabularymobilemanipulation,
      title={Language-Conditioned Open-Vocabulary Mobile Manipulation with Pretrained Models}, 
      author={Shen Tan and Dong Zhou and Xiangyu Shao and Junqiao Wang and Guanghui Sun},
      year={2025},
      eprint={2507.17379},
      archivePrefix={arXiv},
      primaryClass={cs.RO},
      url={https://arxiv.org/abs/2507.17379}, 
}
